\documentclass[lettersize,journal]{IEEEtran}
\usepackage{array}
\usepackage{amssymb}

\usepackage{graphicx}
\usepackage{cite}
\usepackage{subfigure,color}

\usepackage{multirow,amsmath}

\newcommand{\highlight}[1]{\textcolor{black}{#1}}
\newcommand{\etal}{\textit{et al.}}
\newcommand{\mathbbm}[1]{\text{\usefont{U}{bbm}{m}{n}#1}}

\usepackage{pdfrender}
\newcommand*{\boldcheckmark}{%
  \textpdfrender{
    TextRenderingMode=FillStroke,
    LineWidth=.5pt, 
  }{\checkmark}%
}

\begin{document}


\title{TextDCT: Arbitrary-Shaped Text Detection via Discrete Cosine Transform Mask}

\author{Yuchen~Su,
        Zhiwen~Shao,
        Yong~Zhou,
        Fanrong~Meng,
        Hancheng~Zhu,
        Bing~Liu,
        and~Rui~Yao 
\thanks{Manuscript received January, 2022. This work was supported in part by the National Natural Science Foundation of China under Grant 62106268, and in part by the High-Level Talent Program for Innovation and Entrepreneurship (ShuangChuang Doctor) of Jiangsu Province under Grant JSSCBS20211220. It was also supported in part by the National Natural Science Foundation of China under Grants 62101555 and 62172417, in part by the Natural Science Foundation of Jiangsu Province under Grants BK20201346 and BK20210488, in part by the Six Talent Peaks Project in Jiangsu Province under Grant 2015-DZXX-010, and in part by the Fundamental Research Funds for the Central Universities under Grant 2021QN1072. (Yuchen~Su and Zhiwen~Shao contributed equally to this work. Corresponding author: Zhiwen~Shao.)}%
\thanks{Y. Su, Z. Shao, Y. Zhou, F. Meng, H. Zhu, B. Liu, and R. Yao are with the School of Computer Science and Technology, China University of Mining and Technology, Xuzhou 221116, China, and also with the Engineering Research Center of Mine Digitization, Ministry of Education of the People’s Republic of China, Xuzhou 221116, China (e-mail: \{yuchen\_su; zhiwen\_shao; yzhou; mengfr; zhuhancheng; liubing; ruiyao\}@cumt.edu.cn).}
}

\markboth{IEEE Transactions on Multimedia,~Vol.~X, No.~X, X}%
{Shell \MakeLowercase{\textit{et al.}}: Bare Demo of IEEEtran.cls for IEEE Journals}

\maketitle

\begin{abstract}
Arbitrary-shaped scene text detection is a challenging task due to the variety of text changes in font, size, color, and orientation.
Most existing regression based methods resort to regress the masks or contour points of text regions to model the text instances. However, regressing the complete masks requires high training complexity, and contour points are not sufficient to capture the details of highly curved texts.
To tackle the above limitations, we propose a novel light-weight anchor-free text detection framework called TextDCT, which adopts the discrete cosine transform (DCT) to encode the text masks as compact vectors.
Further, considering the imbalanced number of training samples among pyramid layers, we only employ a single-level head for top-down prediction. 
To model the multi-scale texts 
in a
single-level head, we introduce a novel positive sampling strategy by treating the shrunk text region as positive samples, and design a feature awareness module (FAM) for spatial-awareness and scale-awareness by fusing rich contextual information and focusing on more significant features.
Moreover, we propose a segmented non-maximum suppression (S-NMS) method that can filter low-quality mask regressions.
Extensive experiments are conducted on four challenging datasets, which demonstrate our TextDCT obtains competitive performance on both accuracy and efficiency.
Specifically, TextDCT achieves F-measure of 85.1 at 17.2 frames per second (FPS) and F-measure of 84.9 at 15.1 FPS for CTW1500 and Total-Text datasets, respectively. 


\end{abstract}


\IEEEpeerreviewmaketitle

\begin{figure}[!ht]
    \centering
    \subfigure[TextRay~\cite{wang2020textray}]
    {
        \includegraphics[width=0.22\textwidth,height=3cm]{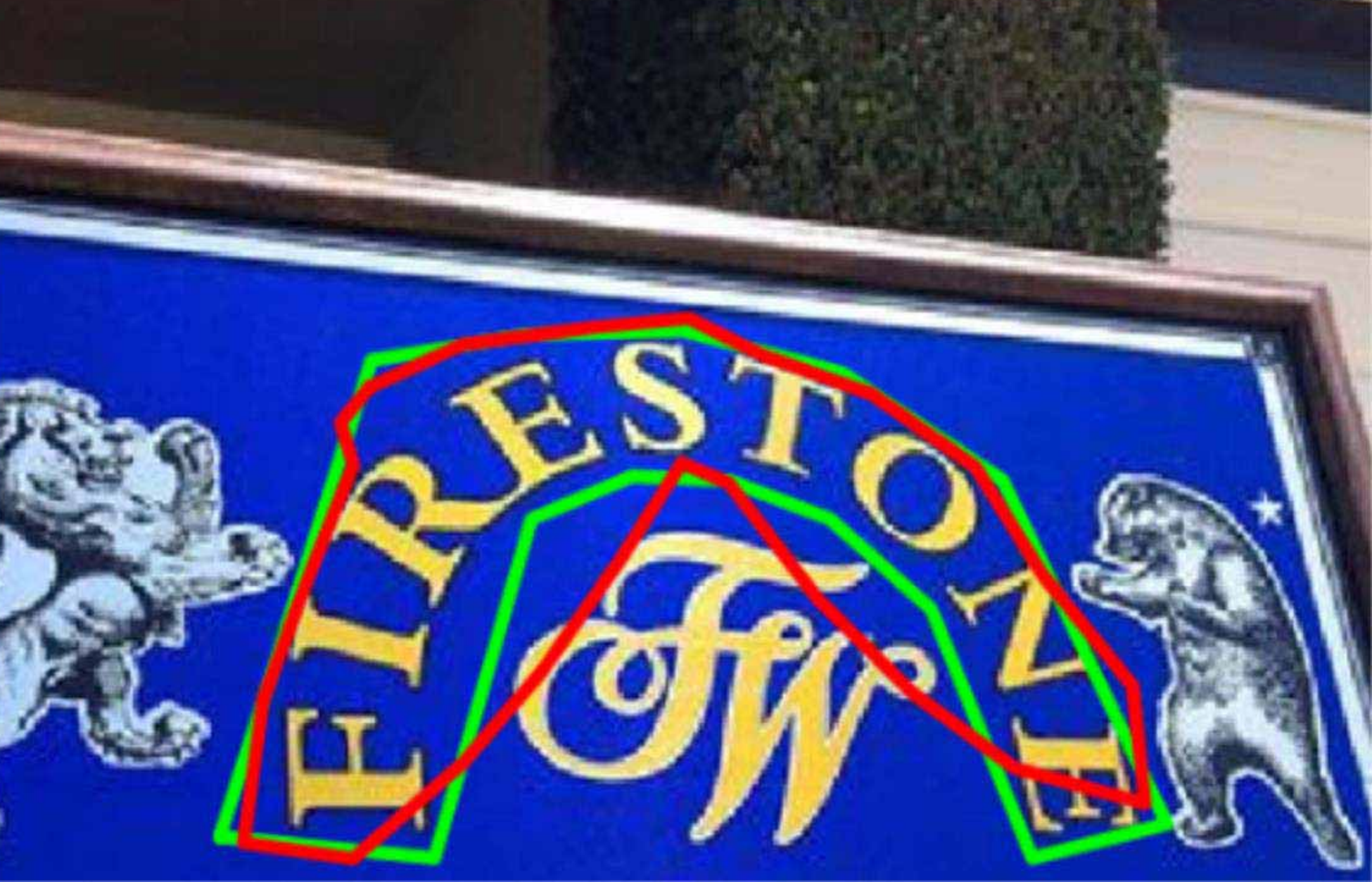}
        \label{fig1a}
    }
    \subfigure[\textbf{TextDCT}]
    {
        \includegraphics[width=0.22\textwidth,height=3cm]{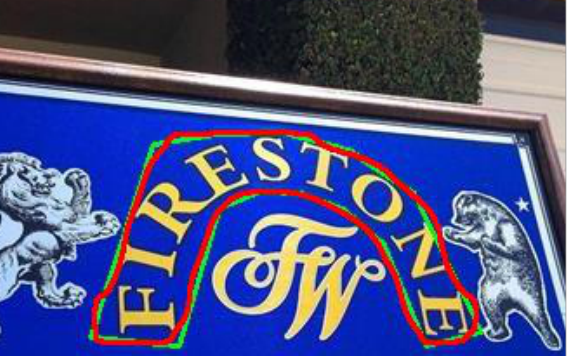}
        \label{fig1b}
    }
    \caption{Comparison of the text detection results of an example image for TextRay\cite{wang2020textray} and our TextDCT. Ground-truth contours are represented in green, and the detection results are represented in red. It can be seen that TextRay fails to model highly curved texts.}
    \label{fig1}
\end{figure}

\section{Introduction}
\IEEEPARstart{O}{ver} the past few years, scene text detection has attracted much attention in the computer vision community, due to its applications in many fields like scene understanding\cite{yi2014scene},  product search\cite{xiong2016text}, and autonomous driving\cite{hong2019textplace}. Benefited from the development of deep learning, scene text detection has made great progress~\highlight{\cite{liao2018textboxes++,dai2019deep, zhang2020opmp,wang2020r,dai2021accurate,xing2021boundary,dai2021comprehensive}}. However, due to unconstrained text variations in font, size, color, and orientation, arbitrary-shaped scene text detection remains a challenge. 



Current scene text detection methods based on deep learning can be divided into two categories: regression based approaches\cite{2019MSR,2019Look,2020Deep} and segmentation based approaches\cite{wang2019Shape,2020Learning,liao2020Real,2019Detecting,xie2019scene}. 
Due to the prediction for each pixel, segmentation based approaches do not need to explicitly process complex curved texts.
However, such approaches are sensitive to noises, so they usually depend on the pre-training on a large dataset.
Besides, pixel-level processing significantly increases the computational cost and the post-processing steps are typically very complicated. In contrast, regression based methods are often more concise and are easier to train.
However, there are still two main problems unresolved for regression based methods. 

On one hand, designing a compact text mask representation that can fit diverse geometry variances of arbitrary-shaped text instances is challenging. Because of the high complexity of directly regressing arbitrary-shaped text masks, most of the existing regression based methods regress contour point sequences of texts. However, point sequences are not sufficient to capture the details of highly curved texts, in which the represented text contour is usually unsmooth, as shown in Fig.~\ref{fig1a}.



On the other hand, state-of-the-art regression based methods rely heavily on the divide-and-conquer strategy in feature pyramid networks (FPN)\cite{lin2017feature} to regress multi-scale texts. However, all training samples are subject to the same supervision, leading to an imbalanced supervision issue among different pyramid layers, especially for single-stage detectors. Specifically, the number of training samples in P3 layer is $256$ times of that in P7 layer of FPN. Thus, connecting multi-level heads together will cause extremely imbalanced learning among samples on different layers, as the shallow layers like P3 receive much more supervision than deeper layers.



To tackle these two problems, we propose a novel arbitrary-shaped scene text detection framework, namely TextDCT. First, inspired by the recent instance segmentation work \cite{shen2021dct}, we model the high-resolution text instance mask in the frequency domain instead of the spatial domain via discrete cosine transform (DCT), and keep its low-frequency components as the mask representation, which has low training complexity. Due to the energy concentration characteristics of DCT, in which most of the natural signal energy is concentrated in the low-frequency components, so the mask representation by this transformation has a high quality.

\begin{figure*}
\centering
\includegraphics[width=0.82\textwidth]{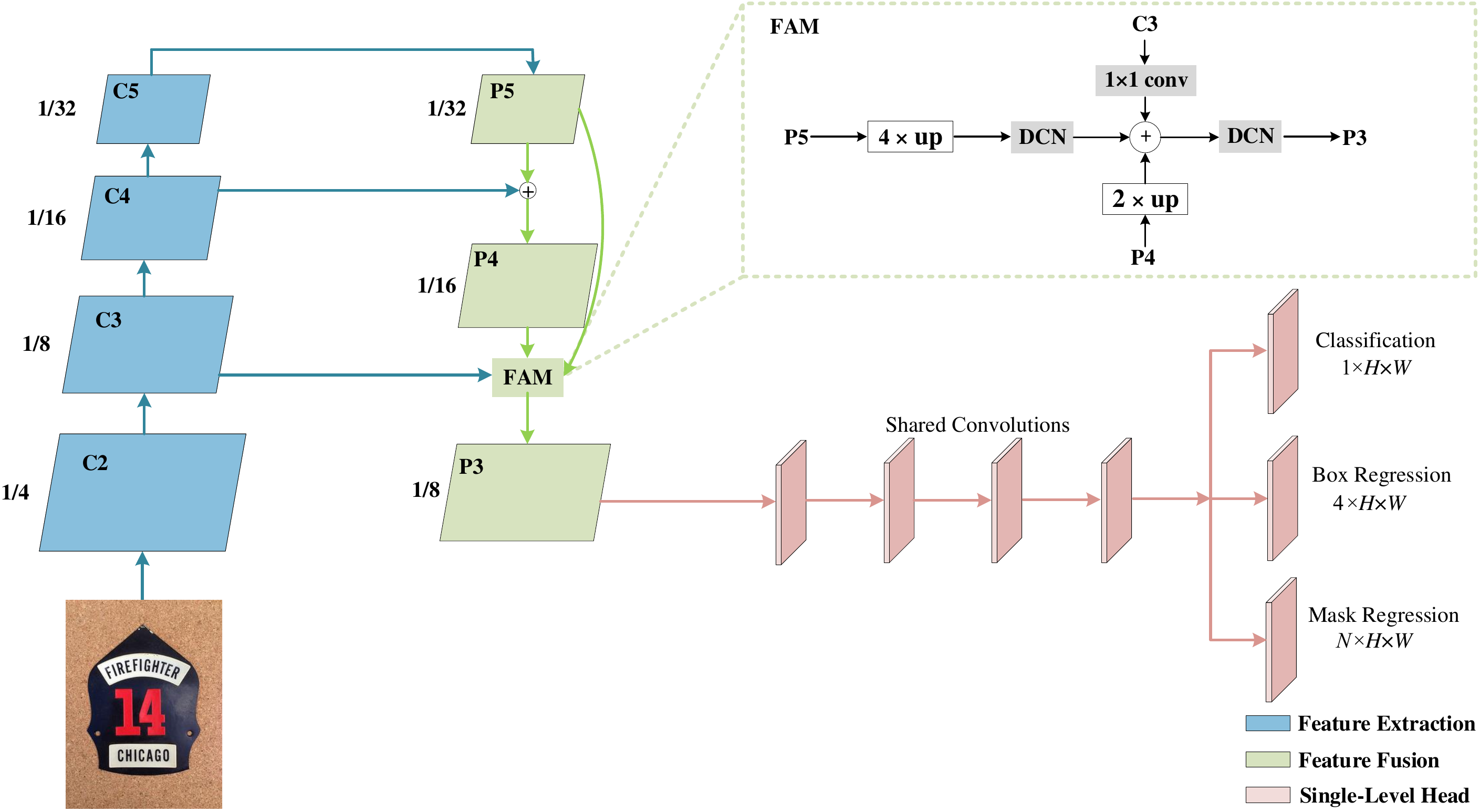}
\caption{The overview of our TextDCT, which is composed of three modules: (a) feature extraction as the backbone, (b) feature fusion via feature pyramid with FAM, and (c) single-level head for three-branch joint optimization. The classification branch and the box regression branch are used for predicting text kernels and text boxes, and the mask regression branch is used for predicting the mask vectors transformed by DCT. Here ``$1/k$'' denotes $1/k$ of the input image size, DCN denotes deformable convolution, ``+'' denotes element-wise sum of feature maps, and ``$k \times$ up'' denotes up-scaling the feature map with a factor $k$.  
}
\label{fig2}
\end{figure*}

 
 
 
Moreover, although recent research efforts\cite{wu2021progressive,chen2021disentangle} can mitigate the imbalanced supervision problem by independent supervision on different pyramid layers, they do not go beyond the divide-and-conquer, which makes its structure more complex in single-stage detectors. Intuitively, using only a single-level head can solve this problem and the model is much simpler. However, directly replacing multi-level heads with a single-level head can lead to a dramatic model performance drop.
The first reason is that the single-level head is not scale-aware and spatial-aware, in which the receptive field of single-level feature cannot match large range of text scales. Since arbitrary-shaped texts usually appear in vastly different fonts, rotations, and shapes, the detection performance for multi-scale text instances using a single-level head is limited. The second one is that a positive sampling strategy suitable for single-level prediction is required. If using the popular ``center sampling'' positive sampling strategy\cite{tian2020fcos}, the masks of different text instances whose center points are in the same region are easily confused.
 
 
Considering the above issues, we design a feature awareness module (FAM) to achieve spatial-awareness and scale-awareness by fusing rich contextual information, capturing larger receptive field and focusing on more significant features. We also introduce a text kernel sampling (TKS) strategy for the single-level prediction by treating the shrunk text kernel region as the positive samples, which can adaptively adjust the number of positive samples to balance text regression at different scales. 
Besides, 
based on the separation of text kernels from each other, we propose segmented non-maximum suppression (S-NMS) to effectively suppress false positives, especially for long text instances.

Our TextDCT is a single-shot, anchor-free, and light-weighted framework that performs joint optimization of text/non-text kernel classification, text bounding box regression, and text mask regression. The main contributions of this work are summarized as follows:
\begin{itemize}
    \item We propose a novel scene text detection framework by employing DCT to represent text masks, which can accurately approximate arbitrary-shaped text instances while having low training complexity.
    \item We design a single-level head for top-down text prediction, with a feature awareness module, a text kernel sampling strategy, and a segmented non-maximum suppression method. These strategies are beneficial for avoiding imbalanced supervision, processing multi-scale text variations, and suppressing false positives.
    \item Extensive experiments demonstrate that our TextDCT achieves competitive performance on both accuracy and efficiency. Particularly, TextDCT obtains F-measure of $85.1$ at $17.2$ frames per second (FPS) and F-measure of $84.9$ at $15.1$ FPS for CTW1500 and Total-Text datasets, respectively.
    
     
\end{itemize}


\section{Related Work}
Current scene text detection methods can be divided into two categories: segmented based scene text detection and regression based scene text detection. 
Besides, we discuss some current false positive suppressing methods.

\subsection {Segmentation Based Scene Text Detection}
Inspired by semantic segmentation methods\cite{long2015fully,ronneberger2015u}, some works regard the arbitrary-shaped text detection as a segmentation problem, which represents complex text instances with pixel-level prediction and rebuilds text instances through specific post-processing. Pixellink\cite{deng2018pixellink} first predicted the linkage relationship between pixels, then extracted the text bounding boxes by separating the links belonging to different text instances. TextSnake\cite{long2018textsnake} treated text instances as a sequence of overlapping disks, and predicted text area, text center line, and some geometric attributes of disks to rebuild the text instances. To effectively split the close text instances, PSENet\cite{wang2019Shape} detected different scale kernels in each text instance, and adopted a progressive scaling method to gradually expand the text kernels to obtain the final detections. Besides, TextField\cite{xu2019textfield} first learnt a direction field containing both text mask and relative position information away from text boundary, then linked neighbor pixels to generate candidate text instances. Tian \etal\cite{tian2019learning}  assumed each text instance as a cluster and performed a two-step clustering strategy to segment dense text instances, where pixels of the same text usually lie in the same cluster. CRAFT\cite{baek2019character} detected character-level text regions by using two-dimensional Gaussian segmentation labels and exploring affinity between characters. DBNet\cite{liao2020Real} introduced a differentiable binarization processing module that gives a high threshold for text boundaries to distinguish adjacent texts. Moreover, TextBPN\cite{zhang2021adaptive} first obtained boundary proposals based on the distance field map and classification map, then deformed the boundary proposals into more accurate text boundaries by the adaptive boundary deformation module.

Most of these methods can adapt to curved texts, but are sensitive to text-like background noises and holes, resulting in false positive cases. In contrast, our TextDCT can deal with false positives well by the S-NMS strategy.

\subsection {Regression Based Scene Text Detection}
Regression based methods typically rely on the object detection pipeline with bounding box regression, which is usually easier to train than segmentation based methods. CTPN\cite{tian2016detecting} first used a modified pipeline of Faster R-CNN\cite{ren2015faster} to detect a set of partial text components with a fixed-size width, then connected them within different instances. 
RRPN\cite{ma2018arbitrary} also used a modified pipeline of Faster R-CNN, which adopted rotated proposals to detect multi-oriented texts. TextBoxes\cite{liao2017textboxes}  modified the shapes of convolutional filters and increased the proportion of default boxes of SSD\cite{liu2016ssd} to adapt to the aspect ratio of texts. TextBoxes++\cite{liao2018textboxes++} extended TextBoxes by applying quadrilateral regression to effectively detect multi-oriented text. EAST\cite{zhou2017east} adopted an anchor-free single-stage detection framework to directly regress the offsets between points within text instances and the corresponding bounding boxes or four corner points. 
MOST\cite{he2021most} proposed a text feature alignment module (TFAM) and a position-aware
non-maximum suppression (PA-NMS) module to achieve accurate detection of long texts. However, the above regression based text representations are horizontal or oriented rectangles, which have limited capacity to model irregular texts. 

Recently, several works~\cite{liu2019curved,2019Look,2020Deep,wang2020textray,liu2020abcnet,dai2021progressive,zhu2021fourier} focused on irregular text detection. CTD-TLOC\cite{liu2019curved} regressed the offsets between the top-left point of the bounding box and the key points on text contours, then smoothed the offsets with a recurrent neural network (RNN).
LOMO\cite{2019Look} introduced a shape representation module that uses center lines, text regions, and border offsets to represent texts, then proposed an iterative refinement module to regress long texts. DRRG\cite{2020Deep} treated text instances as a series of combinations of small rectangular components and introduced a graph convolutional network to learn the linkage relationships of the text components. TextRay\cite{wang2020textray} formulated the text contours in the polar system, and regressed the distance from the polar coordinate to the point where the emitted N-rays intersect with the text boundary. ABCNet\cite{liu2020abcnet} adopted an anchor-free network to predict a series of points, then used Bernstein polynomial to transform these points into Bezier curves that fit the text contours. PCR\cite{dai2021progressive} proposed the contour location mechanism (CLM) to regress text contours progressively. FCENet\cite{zhu2021fourier} treated text contours as periodic functions, and used the discrete Fourier transform (DFT) to convert text contours to Fourier eigenvectors. 
Unlike FCENet using DFT to encode text contours, our method adopts DCT to encode text masks as compact vectors. 


\subsection {False Positive Suppression}
To suppress false positives, SPCNet\cite{xie2019scene} designed a text context module (TCM) and a re-score mechanism to improve the accuracy of the scores of tilted texts. ContourNet\cite{wang2020contournet} designed a local orthogonal texture-aware module (LOTM), which considers the local texture information in two orthogonal directions simultaneously. TextRay\cite{wang2020textray} designed a central-weighted training strategy to give beneficial gradients for long text instances. ABCNet\cite{liu2020abcnet} used the center-ness branch to get better text bounding boxes. TextFuseNet\cite{ye2021scene} fused global-level, char-level and word-level features of texts to strengthen the capability of distinguishing texts and non-texts.
PCR\cite{dai2021progressive} designed a contour localization mechanism to re-score the localized contours. However, these methods increase the training complexity or the post-processing time, which brings additional computational burden. In this paper, we propose the S-NMS to suppress false positives in long texts  
with low overhead.

\begin{figure*}
\includegraphics[width=\textwidth]{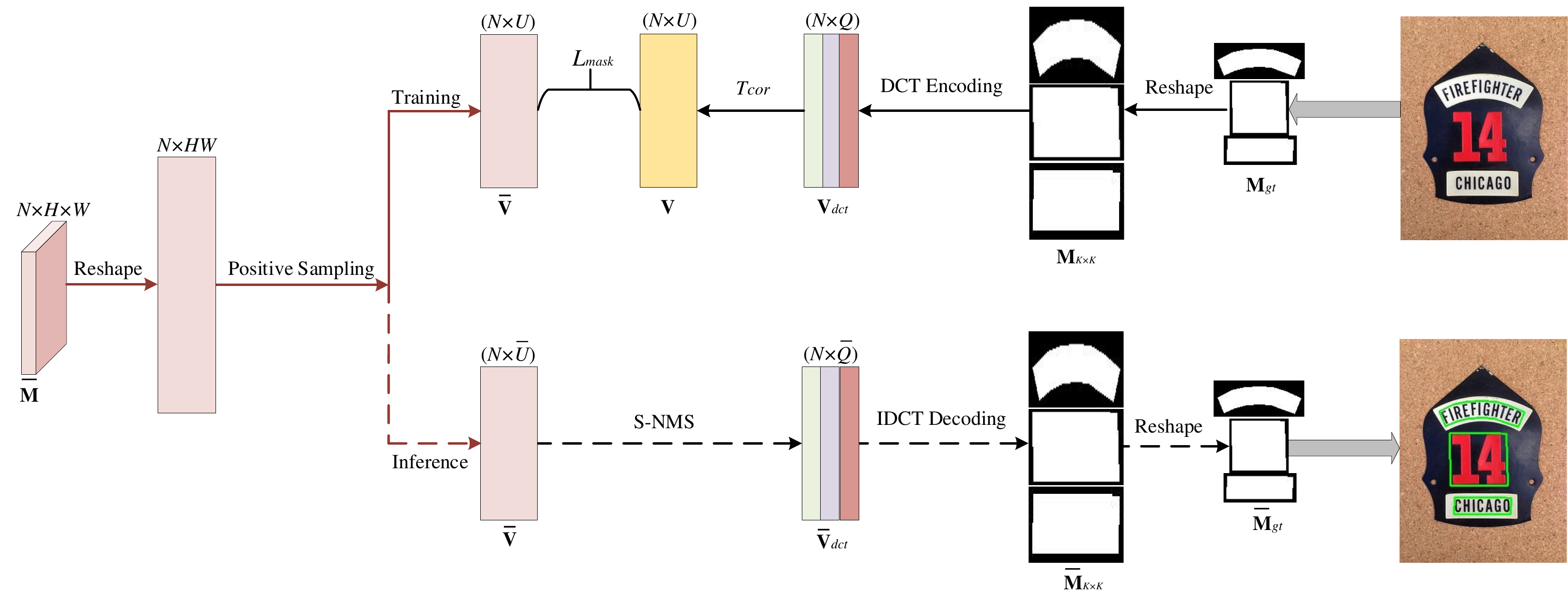}
\caption{The learning of DCT mask representation. $\bar{\mathbf{M}}$ is the output of the mask regression branch in Fig.~\ref{fig2}, and $\bar{\mathbf{V}}$ are predicted mask vectors corresponding to the positive sample points in $\bar{\mathbf{M}}$. ${L}_{mask}$ is the loss for mask regression task. $T_{cor}$ refers to the one-to-one correspondence between the frequency-domain ground-truth and each positive sample point. Both $M_{K \times K}$ and $\bar{M}_{K \times K}$ have a uniform size, which are beneficial for DCT encoding and IDCT decoding. $\bar{M}_{gt}$ and $M_{gt}$ denote the predicted text masks and the ground-truth text masks, respectively. In the shown example image, the number of text instances is $Q=3$.}
\label{fig3}
\end{figure*}

\section{Arbitrary-Shaped Text Detection via Discrete Cosine Transform Mask}

\subsection{Overview}
Our TextDCT is a single-shot, anchor-free detection framework that can fit arbitrary-shaped texts. 
As shown in Fig.~\ref{fig2}, this framework mainly contains three parts: feature extraction, feature fusion, and three-branch joint optimization based on the single-level head. In the feature extraction module, we utilize ResNet50\cite{he2016deep} as the backbone to generate shared feature maps with different receptive fields. We first adopt FPN\cite{lin2017feature} to generate fused features with strong representation, and then add an efficient feature awareness module (FAM) between P4 and P3 of FPN to achieve spatial-awareness and scale-awareness for the single-level head. 
In the three-branch joint optimization module, we share the first four convolutions 
to improve the correlation between the classification and regression tasks, and then use three convolutions to predict the probability of each position corresponding to the text kernel, the regression value of each position corresponding to the text bounding box, and the text mask, respectively.

\subsection {DCT Mask Representation}
Most of the existing contour points modeling methods have a limited capacity to represent highly curved or long texts, while the directly regressed high-resolution masks contain redundant information since the discriminative pixels are mainly distributed along the text boundaries. To obtain low-complexity and high-quality text mask representations, we encode the text masks 
into 
compact vectors by DCT. 
As shown in Fig.~\ref{fig3}, the output $\bar{\mathbf{M}}\in{\mathbb{R}^{N \times H \times W}}$ of the mask regression branch in Fig.~\ref{fig2} is first reshaped to be the size of $N \times HW$, where $H$ and $W$ are the height and width of $\bar{\mathbf{M}}$, and $N$ denotes the vector dimension at each point in $\bar{\mathbf{M}}$. 
Then, we regress only the mask vectors $\mathbf{\bar{V}}\in{\mathbb{R}^{N \times U}}$ corresponding to the positive samples in $\bar{\mathbf{M}}$, where $U$ is the number of positive sample points.  

During training, the generation process of the ground-truth mask vector corresponding to each point consists of three steps. Firstly, we reshape the ground-truth mask $\mathbf{M}^j_{gt}$ of the $j$-th text instance to be $\mathbf{M}_{K \times K}^j\in{\mathbb{R}^{K \times K}}$, where $K \times K$ is the uniform mask size, $j=1,\cdots, Q$, and $Q$ is the number of text instances in the input image. 

Secondly, we employ two-dimensional DCT to encode $\mathbf{M}_{K \times K}^j$ into the frequency domain:
\begin{equation}\label{dct}
\begin{aligned}
\mathbf{M}_{d c t}^j&(u, v)=\frac{2}{K} C(u) C(v)[\sum_{x=0}^{K-1} \sum_{y=0}^{K-1}\\ &
\mathbf{M}_{K \times K}^j(x, y)\cos \frac{(2 x+1) u \pi}{2 K} \cos \frac{(2 y+1) v \pi}{2 K}],
\end{aligned}
\end{equation}
where $C(w) = 1/ \sqrt{2}$ for $w = 0$, and $C(w) = 1$ otherwise.
Due to the energy concentration characteristics of DCT, in which most of the natural signal energy is concentrated in the low-frequency components, the compact vector ${V}^j_{dct}\in{\mathbb{R}^N}$ can be sampled from the first $N$-dimensional vector of the $\mathbf{M}_{dct}^j$ in a ``zigzag''. 

Thirdly, we assign ${V}^j_{dct}$ to ${V}^i\in{\mathbb{R}^N}$ by $T_{cor}$, where $i=1,\cdots, U$. $T_{cor}$ is a label assignment strategy that matches the frequency-domain labels to corresponding positive sample points.

During inference, there are three steps for the prediction of text instance masks.
Firstly, we employ a segmented non-maximum suppression (S-NMS) strategy to obtain $\bar{\mathbf{V}}_{dct}$ from the prediction $\bar{\mathbf{V}}$,  which will be elaborated in Sec.~\ref{ssec:snms}. Secondly, we expand $\bar{V}^j_{dct}$ to $K^2$-dimensions by backward complementing 0, and then transform $\bar{V}^j_{dct}$ to 
$\mathbf{\bar{M}}_{dct}^j\in{\mathbb{R}^{K \times K}}$ by reshaping the size of $\bar{V}^j$ to $K \times K$. Thirdly, $\mathbf{\bar{M}}_{K \times K}^j$ can be generated by two-dimensional inverse discrete cosine transform (IDCT):
\begin{equation}\label{idct}
\begin{aligned}
\mathbf{\bar{M}}^j_{K \times K}&(x, y)=\frac{2}{K} [\sum_{x=u}^{K-1} \sum_{v=0}^{K-1}C(u) C(v)\\ & 
\mathbf{\bar{M}}_{d c t}^j(u, v) \cos \frac{(2 x+1) u \pi}{2 K} \cos \frac{(2 y+1) v \pi}{2 K}].
\end{aligned}
\end{equation}
Note that the computational cost of DCT and IDCT are negligible, since the complexity of computation through fast cosine transform (FCT)\cite{haque1985two} is only $O(nlogn)$. 

\subsection {Single-Level Prediction}

State-of-the-art regression based text detectors often adopt the divide-and-conquer strategy, which introduces an imbalanced supervision issue and causes a more complex structure in single-stage detectors. To suppress this issue, our TextDCT framework uses a single-level prediction structure. Two keys to the single-level prediction lie in how to design the input feature of the head so that the head is scale-aware and spatial-aware, and how to design a positive sample assignment strategy.

\subsubsection{Feature Awareness Module}

As shown in Fig.~\ref{fig2}, we use the pyramid feature P3 as the input of the single-level head. Due to the diversity of text scale variations, the receptive field of P3 can only cover a limited scale range. As shown in Fig.~\ref{fig4a}, extreme long texts cannot be accurately regressed using P3 as the head's input, so we design a feature awareness module (FAM).


The structure of FAM is shown in Fig.~\ref{fig2}, which consists of a skip connection and two deformable convolutions. Specifically, we first add a skip connection from P5 to enrich the contextual information of P3. Since direct element-wise sum of P3 and upsampled P5 leads to misaligned contexts of fused features, which may harm the prediction around text boundaries, 
we add a deformable convolution in the skip connection to learn offsets from spatial differences between P3 and unsampled P5, so as to obtain the aligned feature with contextual information. 
Besides, considering deformable convolution has the ability to focus on salient regions from various text appearances, we add a deformable convolution after the fused feature to capture more significant features. By obtaining features with rich contextual information and adaptively adjusting the receptive fields to achieve spatial-awareness and scale-awareness for the single-level head, our proposed FAM can suppress the limitations of the divide-and-conquer strategy.


\subsubsection{Positive Sampling Strategy}
\begin{figure}[t]
    \centering
    \subfigure[Baseline]
    {
        \includegraphics[width=0.22\textwidth,height=1.8cm]{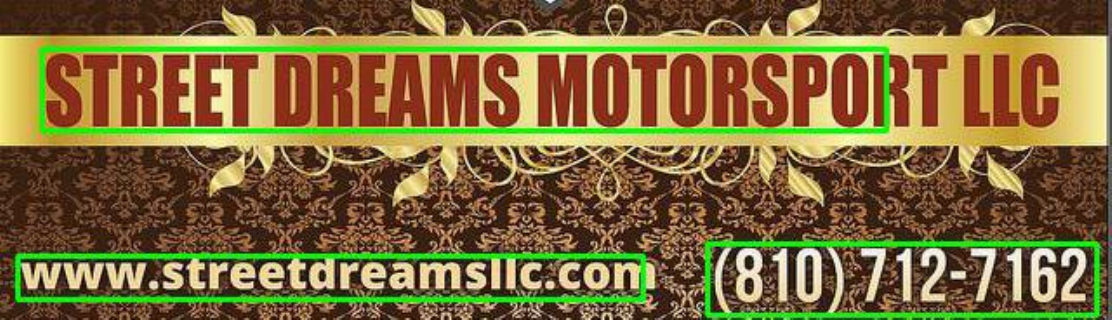}
        \label{fig4a}
    }
    \subfigure[\textbf{Baseline+FAM}]
    {
        \includegraphics[width=0.22\textwidth,height=1.8cm]{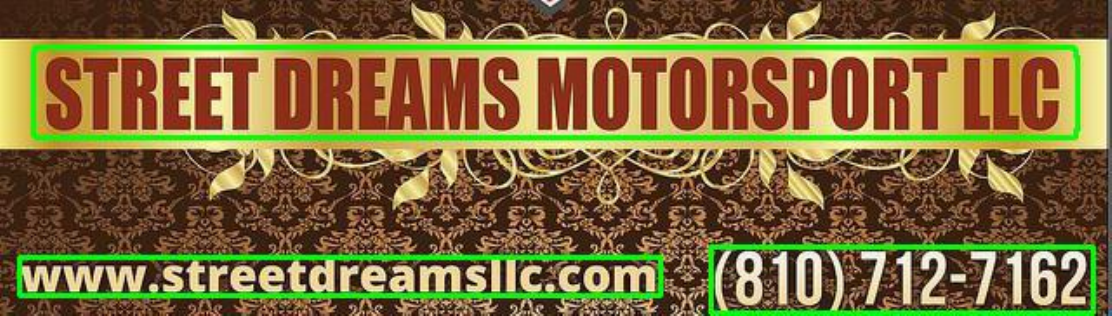}
        \label{fig4b}
    }
    \subfigure[Center sampling]
    {
        \includegraphics[width=0.22\textwidth,height=1.8cm]{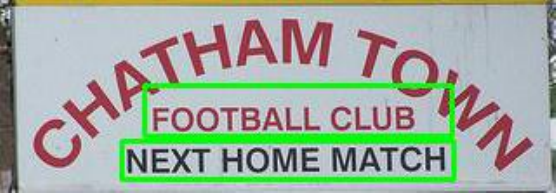}
        \label{fig4c}
    }
    \subfigure[\textbf{TKS}]
    {
        \includegraphics[width=0.22\textwidth,height=1.8cm]{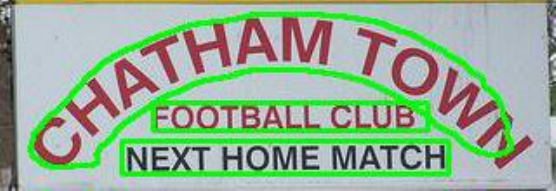}
        \label{fig4d}
    }
    \caption{Illustration of the superiority of our proposed FAM and TKS. The comparison between (a) and (b) shows that adding FAM can better regress long texts, and the comparison of (c) and (d) shows that TKS is more effective for single-level prediction.}
    \label{fig4}
\end{figure}

The definition of positive samples is crucial for text detection, and balanced positive and negative samples are beneficial for achieving accurate classification results.
Most existing anchor-free single-stage detectors adopt the "center sampling" strategy\cite{tian2020fcos} to define positive samples. However, this strategy is not suitable for the case of single-level head, as potentially numerous ambiguous points harm the accuracy of regressing text shapes, as shown in Fig.~\ref{fig4c}. Besides, it is unreasonable to give the same number of positive samples to texts of different sizes. Therefore, we propose a text kernel sampling (TKS) strategy by treating the text kernel as the positive sample region, in which the text kernel can be obtained by Vatti clipping algorithm\cite{vatti1992generic}. Since each point in the text kernel is located inside the text instance, there are few ambiguous samples, and the number of positive samples changes adaptively with the text scales. In this way, we can effectively alleviate the bias between texts with different scales during training. As shown in Fig.~\ref{fig4d}, our proposed TKS can generate more high-quality predictions.

\subsection{Segmented Non-Maximum Suppression}
\label{ssec:snms}

\begin{figure}[t]
    \centering
    \subfigure[]
    {
        \includegraphics[width=0.22\textwidth,height=2.8cm]{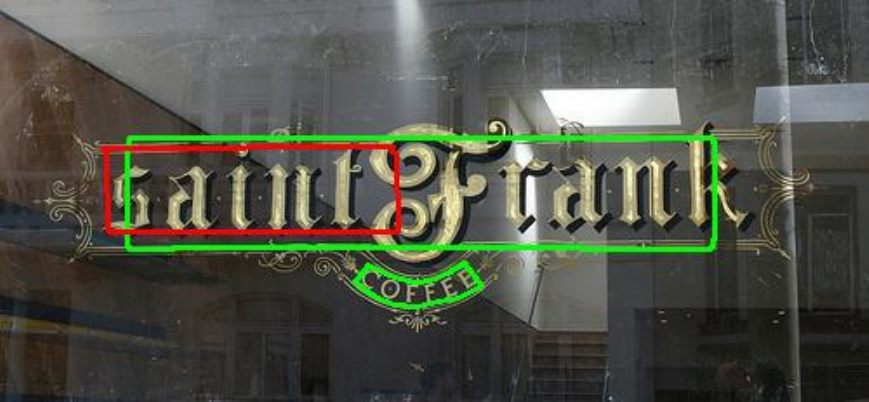}
    }
    \subfigure[]
    {
        \includegraphics[width=0.22\textwidth,height=2.8cm]{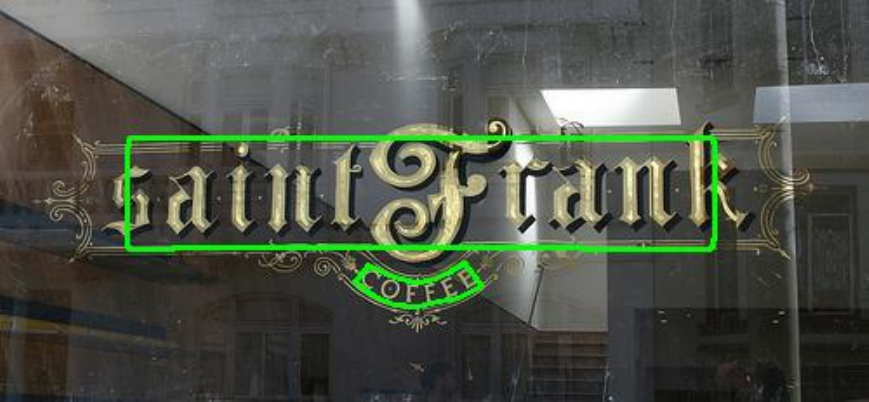}

    }
    \caption{Visualization of false positive suppression strategies. (a) The results of using NMS. (b) The results of using S-NMS. The red contour in (a) represents the false positive detection.}
    \label{fig5}
\end{figure}

As shown in Fig.~\ref{fig2}, a given image goes through our TextDCT, in which the classification scores predicted by the classification branch and the text boxes regressed by the box regression branch can be combined to obtain text box predictions.
Typically, non-maximum suppression (NMS) is used to remove duplicated box predictions. 
However, the points near the text kernel edges of long text instances are generally far from the centers of the texts, making it hard to perceive the comprehensive text shape information accurately. Therefore, the regressed boxes corresponding to these points may be inaccurate and may have a low intersection over union (IoU) with the boxes corresponding to the points near the centers of the texts, which makes it difficult to be filtered by the NMS and thus leads to 
false positives, as illustrated in Fig.~\ref{fig5}. 

TextRay\cite{wang2020textray} and ABCNet\cite{liu2020abcnet} address this problem by using a center-weighted training strategy and adding a center-ness branch, respectively, in which the training complexity is increased. 
Instead, we adopt a segmented non-maximum suppression (S-NMS) strategy. In particular, benefited from the strategy of using text kernels as positive samples, each text kernel can be easily distinguished. 
For each text kernel, we firstly select only the box corresponding to the point with the highest classification score, which allows us to filter out other boxes that may be of low quality directly without considering the IOU threshold. Then, we use NMS to filter the remaining potentially duplicate boxes. 

\begin{figure}
    \centering
    \subfigure[]
    {
        \includegraphics[width=0.22\textwidth,height=2.8cm]{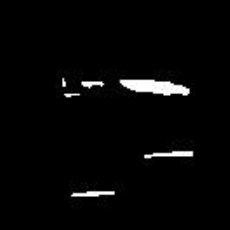}
        \label{fignmsa}
    }
    \subfigure[]
    {
        \includegraphics[width=0.22\textwidth,height=2.8cm]{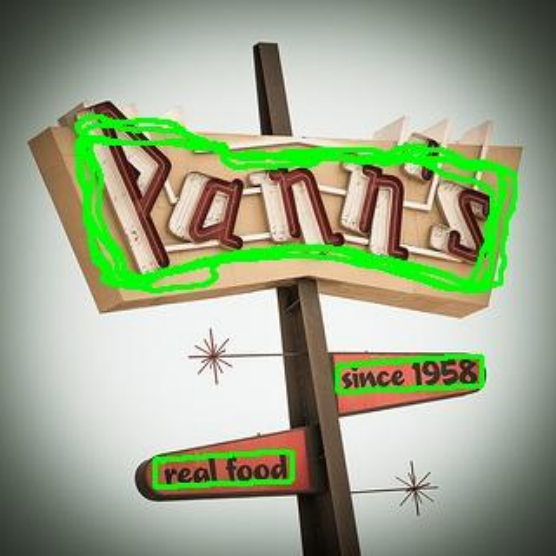}
        \label{fignmsb}
    }
    \caption{Visualization of the effectiveness of the regression based methods. (a) The results of predicted text kernel. (b) The final detection results corresponding to the highest classification scores for all text kernels.}
    \label{fignms}
\end{figure}

Note that, as shown in Fig.~\ref{fignmsa}, there are still duplicate boxes after the first step of the S-NMS because the widespread holes in the text may cause the text kernel to be split into multiple parts. Due to the regression task prefers to focus on the edge parts, as shown in Fig.~\ref{fignmsb}, where most of the points belonging to the same kernel but split into different parts can still regress to the same text instance.  

\subsection{Full Loss Function}
In our TextDCT framework, the full loss function is formulated as
\begin{equation}
\label{eq3}
\mathcal{L}=L_{cls}+\lambda_{1} L_{box}+\lambda_{2} L_{mask},
\end{equation}
where $\lambda_{1}$ and $\lambda_{2}$ are the trade-off factors of the loss function.
$L_{cls}$, $L_{box}$ and $L_{mask}$ denote classification loss, bounding box regression loss and mask vector regression loss of the three branches, respectively.

\begin{equation}\label{dce}
L_{c l s}=1-\frac{2 \sum_{i} P_{k e r}(i) G_{k e r}(i)}{\sum_{i} P_{k e r}(i)+\sum_{i} G_{k e r}(i)},
\end{equation}

\begin{equation}\label{giou}
L_{box}=1-I O U+\frac{|C-(A \cup B)|}{|C|},
\end{equation}
where we choose Dice Loss to optimize $L_{cls}$,  
$G_{k e r}(i)$ and $ P_{k e r}(i)$ denote the $i$-th pixel value in the ground-truth and the prediction of text kernels, respectively. Besides, we use GIoU loss for $L_{box}$ following ABCNet\cite{liu2020abcnet}, 
where A, B are ground truth and predicted bounding boxes, respectively. C is the smallest convex box enclosing both A and B and $I O U=|A \cap B| /|A \cup B|$. 

Our mask vector regression loss is defined as
\begin{equation}
L_{mask}=\mathbbm{1}^{text} \sum_{i}^{N} d_{mask}\left(\bar{v}_{i}, v_{i}\right),
\end{equation}
where $v_{i},\bar{v}_{i}$ represent the $i$-th element in ground-truth and prediction mask vectors, respectively. $\mathbbm{1}^{text}$ is the indicator function for text kernels. $d_{mask}$ is the vector loss, in which we choose smooth-L1 loss for its effectiveness and stability in training.

\subsection{Post-Processing}
The post-processing of our TextDCT model consists of four steps. Firstly, 
the locations of the positive samples are obtained according to the classification threshold $\tau_a$, and then their corresponding predicted text boxes and text mask vectors are obtained. Secondly, the S-NMS is utilized to filter the overlapped text boxes. Thirdly, the text mask vectors corresponding to the remaining text boxes are transformed to 2D text masks by IDCT, and then the spatial resolutions of these text masks are resized to the same sizes of the spatial resolutions of their corresponding text boxes. Finally, these resized text masks are binarized by a threshold $\tau_b$, and then are projected into the complete map with the same spatial resolution as the input image based on the locations of their corresponding text boxes.

\section{Experiments}

\subsection{Datasets and Settings}
\subsubsection{Datasets}
We evaluate our TextDCT on four popular text detection datasets: CTW1500, Total-Text, ICADAR2015, and MLT.
\begin{itemize}
    \item \textbf{CTW1500} \cite{liu2019curved} is a challenging dataset that contains irregular-shaped and multi-oriented texts. It consists of $1,000$ training images and $500$ test images. Texts in this dataset predominantly suffer from blurring, low resolution and perspective distortion, and the text regions are all annotated by $14$ key points.

    \item \textbf{Total-Text} \cite{ch2017total} consists of $1,555$ images ($1,255$ for training and $300$ for testing). It contains horizontal, multi-oriented, and curved texts. Unlike CTW1500, all texts in Total-Text are annotated by world-level polygons with the adaptive number of vertices.
    
    \item \textbf{ICDAR2015} \cite{karatzas2015icdar} consists of $1,000$ natural images for training and $500$ images for testing, including many multi-orientated and street-viewed text instances. The ground-truth of each text is annotated with eight coordinates to enclose the text in a clockwise way.

    \item \textbf{MLT} \cite{nayef2017icdar2017} is proposed on ICDAR 2017 Competition, which involves multi-lingual, multi-script and multi-oriented scene texts. It contains $9,000$ images for training ($7,200$ training images and $1,800$ validation images) and $9,000$ images for testing.
\end{itemize}

\subsubsection{Implementation Details}
The backbone of our TextDCT framework is the pre-trained ResNet50
on ImageNet\cite{krizhevsky2012imagenet}. During training, stochastic gradient descent (SGD) is adopted as an optimizer with a batch size of $2$ for the CTW1500 and Total-Text datasets and $4$ for the MLT and ICDAR2015 datasets. The model is trained up to $120,000$ iterations with the initial learning rate of $0.001$, in which the learning rate is decreased to $0.0001$ at the $90,000$-th iteration. In addition, we employ a pre-trained model trained with $180,000$ iterations on the SynthText\cite{gupta2016synthetic} dataset. Following other methods~\cite{liu2020abcnet,liao2020Real,wang2019efficient}, the text regions labeled as ``DO NOT CARE'' are ignored during training. Our TextDCT is trained on an NVIDIA Tesla V100 GPU. It takes about $10$ training hours on CTW1500 and Total-Text datasets, and takes about $22$ training hours on ICDAR2015 and MLT datasets.

We set $K$ to $64$, $N$ to $300$, $\tau_a$ to $0.9$, $\tau_b$ to $0.35$, and the shrinking rate of the text kernels to $0.5$ for all datasets. Besides, we use the multi-scale training strategy, in which the short side of the input images is set to $640$, $672$, $704$, $736$, $768$, $800$, $832$, $864$, and $896$, while the long side is maintained to $1,024$ on CTW1500 and Total-Text datasets, and is maintained to $1,600$ on ICDAR2015 and MLT datasets. In Eq.~\eqref{eq3}, $\lambda_1=1$, and $\lambda_2=1$.
For data augmentation, we perform random horizontal flip, random crop, color jitter and contrast jitter for input images. In random cropping, we only crop the non-text regions, in which the crop area is larger than half of the original image area. 
During inference, the input images are resized as $1,056\times1,440$, $1,024\times2,000$, $1,500\times2,600$, and $1,600\times2,000$ for CTW1500, Total-Text, ICDAR2015, and MLT, respectively. In the following sections, we omit $\%$ for simplicity in the recall (R), precision (P), and F-measure (F) results.

\subsection{Ablation Study}
In this section, we conduct ablation studies on both Total-Text and CTW1500 datasets to validate main modules in our TextDCT.


\begin{figure}[t]
    \centering
    \subfigure[Baseline]
    {
        \includegraphics[width=0.22\textwidth,]{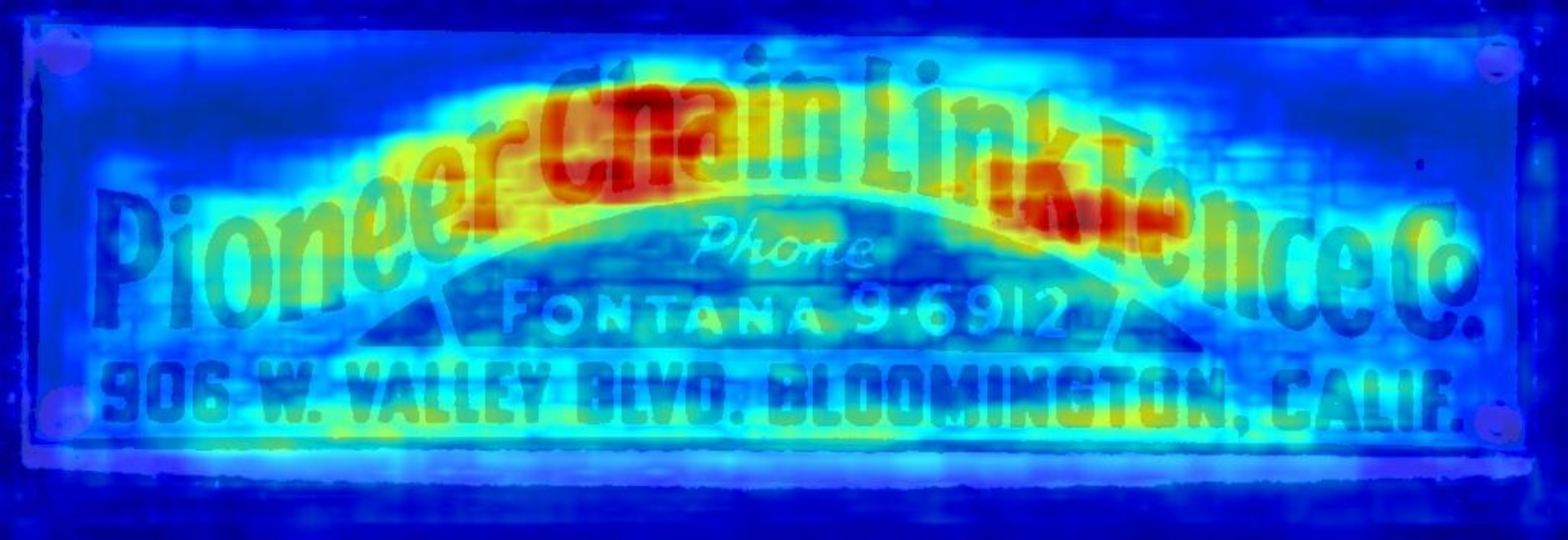}
    }
    \subfigure[\textbf{Baseline+FAM}]
    {
        \includegraphics[width=0.22\textwidth,]{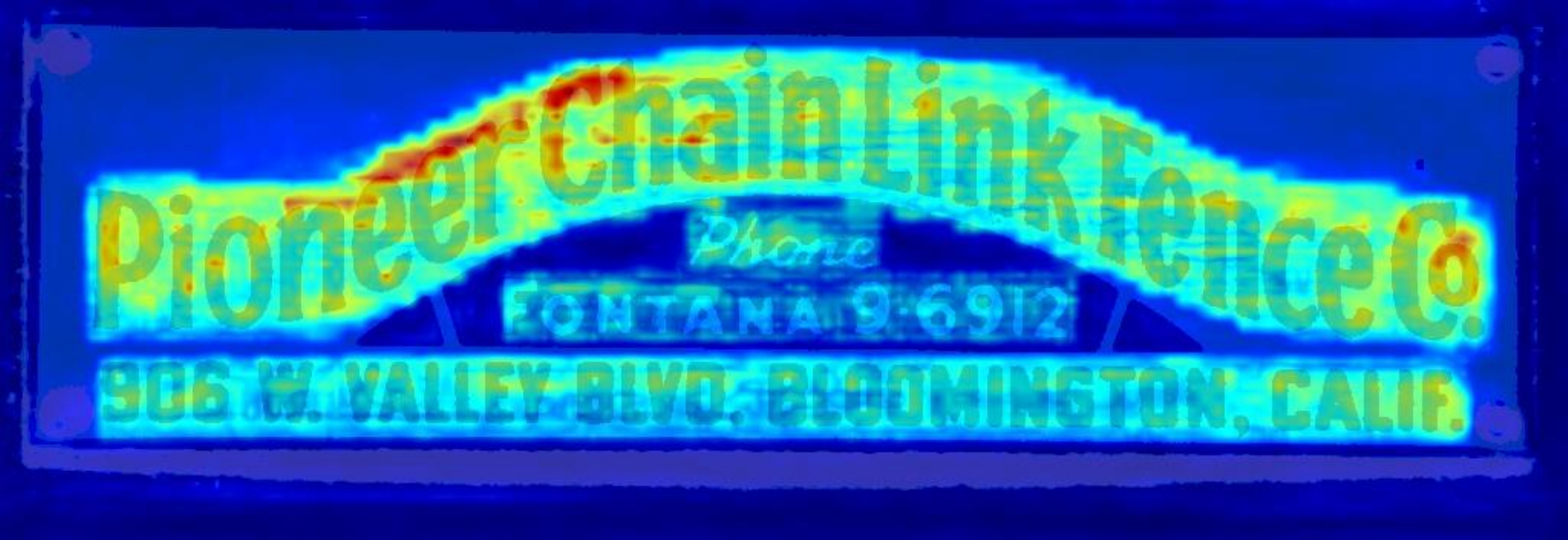}
    }
    \caption{\highlight{
    Visualization of the feature heatmaps of an example image for Baseline and Baseline+FAM. For better viewing, we take channel-wise maximum value at each pixel of P3 to obtain the feature heatmap, and overlay the feature heatmap on the example image.}}
    \label{fig-heat}
\end{figure}

\begin{table}[]
\renewcommand\arraystretch{1.2}
 \centering
 \caption{Performance of using different pyramid layers as head's input on CTW1500.}
 \label{tabel1}
\begin{tabular}{c|ccc} 

\hline
Head's input & Recall        & Precision     & F-measure     \\ \hline
   P5        &  $65.5$           & $78.7$          & $71.5$          \\
   P4        &  $67.8$           & $78.0$          & $72.6$          \\
   \textbf{P3 (Baseline)}        &  $70.7$           & $77.6$          & $74.0$          \\
Muti-level   & $\mathbf{77.4}$ & $\mathbf{81.2}$ & $\mathbf{79.2}$ \\ \hline
\end{tabular}
\end{table}

\subsubsection{Input of the head} 

Table~\ref{tabel1} shows the results of a baseline model using different pyramid feature layers on CTW1500 dataset. The baseline model is TextDCT without FAM and S-NMS, in which the first four convolutions of the head are not shared, and the positive sampling strategy is ``center sampling''\cite{tian2020fcos}. 
Compared to multi-level head inputs, using only a single-level head has a significant performance drop, especially for the recall rate. 
This is mainly because the single-level head without FAM is not scale-aware and spatial-aware, and the ``center sampling'' strategy is not suitable for single-level prediction. 
Although the performance of using multi-level heads is higher than that of single-level head, the imbalanced supervision problem limits its performance. Besides, the F-measure of P3 is $1.4$ and $2.5$ higher than that of P4 and P5, respectively. It demonstrates that P3 makes a good trade-off between semantic information and boundary information compared to P4 and P5. 

\begin{table}[]
\renewcommand\arraystretch{1.2}
 \centering
 \caption{Performance gains of FAM and TKS on curve texts.}
 \label{tabel2}
\begin{tabular}{c|c|ccc} 
\hline
Dataset    & Method         & Recall        & Precision     & {F-measure}     \\ \hline
           & Baseline       & $70.7$          & $77.6$          & $74.0$          \\
\multirow{2}{*}{CTW1500}    & Baseline+FAM & $78.0$          & $81.1$          & $79.5$          \\
           & Baseline+TKS & $76.8$          & $80.1$          & $78.4$          \\
           & \textbf{Baseline+FAM+TKS}        & $\mathbf{80.8}$ & $\mathbf{82.6}$ & $\mathbf{81.7}$ \\ \hline
           & Baseline       & $73.7$      & $79.9$          & $76.7$          \\
\multirow{2}{*}{Total-Text} & Baseline+FAM & $78.0$          & $82.2$          & $80.1$          \\
           & Baseline+TKS & $77.7$          & $81.9$          & $79.7$          \\
           & \textbf{Baseline+FAM+TKS}        & $\mathbf{80.3}$ & $\mathbf{84.0}$ & $\mathbf{82.1}$ \\ \hline
\end{tabular}
\end{table}

\begin{table}[]
 \centering
 \caption{Performance gains of different components in the FAM on CTW1500. SC, DCN1, DCN2 denote the skip connection of P5 to P3 and the two deformable convolutions in the FAM, respectively. DCN3 denotes replacing the $1\times 1$ convolution in FAM with a deformable convolution}.
 \renewcommand\arraystretch{1.2}
 \label{tabelfam}
\begin{tabular}{cccc|ccc}
\hline
SC & DCN1 & DCN2 & DCN3  & R & P & F \\
\hline
-  &- &-  &-   & $76.6$   & $80.6$      & $78.6$      \\
\checkmark  &- &-  &-       & $76.3$   & $81.4$      & $78.8$      \\
\checkmark  &\checkmark &-   &-   & $80.1$   & $82.9$      & $81.5$      \\
-     &- &\checkmark  &-    & $80.4$   & $84.0$      & $82.2$     \\
-     &- &\checkmark  &\checkmark    & $79.8$   & $84.6$      & $82.2$     \\
\boldcheckmark     &\boldcheckmark &\boldcheckmark  &-    & $\mathbf{81.5}$   & $\mathbf{84.7}$      & $\mathbf{83.1}$

\\
\hline
\end{tabular}
\end{table}

\subsubsection{FAM and TKS} 
Table~\ref{tabel2} shows the results of using FAM and TKS over the baseline model on CTW1500 and Total-Text datasets. We can see that the baseline model with FAM achieves the gain of $5.5$ on CTW1500 in terms of F-measure. \highlight{In Fig.~\ref{fig-heat}, the baseline model with FAM has high responses on all the text regions, while the baseline model highlights on a few non-text regions and misses some text regions. This demonstrates that FAM is beneficial for accurately perceiving the overall layout of the texts with different sizes and shapes.}
Besides, using TKS also significantly improves the model performance, with recall and F-measure improved by $6.1$ and $4.4$ on CTW1500, respectively. 
When using both FAM and TKS, the performance is further improved on both CTW1500 and Total-Text datasets.


\subsubsection{Components of FAM}
\label{sssec:fam}

Here we investigate the effectiveness of each component in FAM. FAM consists of a skip connection from P5 to P3 and two deformable convolutions. As shown in Table~\ref{tabelfam}, compared to the model of TextDCT without FAM, further adding a skip connection from P5 to P3 almost has no effect on the performance. 
However, the combination of skip connection and DCN1 improves recall and F-measure by 3.8 and 2.7, respectively, which shows that the features with rich contextual information aligned by deformable convolution are helpful for detecting multi-scale texts. Besides, if only adding DCN2 after P3, the F-measure can reach 82.2, which shows that adaptively adjusting the receptive field to focus on important regions can significantly improve the performance of the single-level prediction model. We also implement a variant by using both DCN2 and DCN3, in which DCN3 denotes replacing the $1\times 1$ convolution in FAM with a deformable convolution. We can observe that the use of DCN3 almost brings no impact on the performance. Therefore, the performance gain by FAM is mainly due to the rich aligned contextual information and the adaptively adjusted receptive field, instead of the deformable convolution itself. When SC, DCN1, and DCN2 are combined together, our TextDCT achieves the best performance.


\subsubsection{DCT mask representation} 
\label{sssec:DCT}
To investigate the effectiveness of the DCT mask representation, 
we implement different variants on CTW1500 dataset, as presented in Table~\ref{tabel4}. IOU refers to the intersection over union between reconstructed mask $\mathbf{\bar{M}}_{gt}$ and ground-truth $\mathbf{M}_{gt}$, which is used to evaluate the quality of the mask representation. 
Without the DCT mask representation, a high dimension makes the model difficult to optimize, leading to the performance degradation. On the other hand, a low resolution leads to significant reconstruction errors, which makes the model difficult to fit complex text shapes accurately.

Firstly, by comparing the $4,096$-dimensional mask vector without DCT and the $300$-dimensional DCT vector under the same resolution of $64 \times 64$, we can see
that the DCT mask representation using lower training complexity can obtain higher recall, precision, and F-measure results.
Secondly, we compare the effect of different resolutions for the $300$-dimensional DCT mask vector. As the resolution increases from $32 \times 32$ to $64 \times 64$, the IOU increases from $95.0$ to $97.4$, and the F-measure gains $0.8$ improvement. When the resolution is further increased to $128 \times 128$, the results are almost unchanged. Therefore, we set the resolution to $64 \times 64$ in our TextDCT.
Finally, we compare different dimensions of the DCT vectors. When the dimension is increased from $300$ to $500$, the IOU only grows $0.2$ and the F-measure remains unchanged at $83.1$. The is due to the energy concentration characteristics of DCT, in which the former dimension is more important than the latter dimension for reconstructed masks. 

\begin{table}[]
 \centering
 \renewcommand\arraystretch{1.2}
 \caption{Ablation study of the resolution of mask representation and the dimension of mask vector on CTW1500. ``*'' means without DCT.} \label{tabel4}
\begin{tabular}{cc|cccc}
\hline
Resolution & Dim  & Recall        & \multicolumn{1}{c}{Precision} & F-measure     & IOU           \\ \hline
$32 \times 32$    & $1,024$* & $79.7$          & $85.0$                          & $82.3$          & $95.6$          \\
$32 \times 32$    & $300$  & $80.0$          & $84.6$                          & $82.3$          & $95.0$          \\
$64 \times 64$    & $4,096$* & $77.0$          & $84.3$                          & $80.5$          & $\mathbf{97.8}$          \\
$64 \times 64$   & $100$  & $81.0$          & $83.6$                          & $82.3$          & $94.2$\\          
$\mathbf{64 \times 64}$    & $\mathbf{300}$  & $\mathbf{81.5}$          & $84.7$                 & $\mathbf{83.1}$          & $97.4$          \\
$64 \times 64$    & $500$  & $81.0$ & $85.3$                          & $\mathbf{83.1}$ & $97.6$ \\
$128 \times 128$  & $300$  & $80.5$          & $\mathbf{85.6}$                          & $83.0$          & $97.4$          \\ \hline

\end{tabular}
\end{table}

\begin{table}[]
 \centering
 \renewcommand\arraystretch{1.2}
 \caption{Performance of sharing the first four convolutions in the head on CTW1500.``gap'' refers to the absolute value of the difference between the classification score and the location score.}
 \label{sc}
\begin{tabular}{c|cccc}
\hline
Shared                    & R & P & F &gap\\ \hline
None                      & $80.6$   & $84.7$      & $82.6$ & $\mathbf{0.22}$    \\
Regression(Box+Mask)      & $80.6$   & $\mathbf{84.9}$      & $82.7$  & $\mathbf{0.22}$    \\
\textbf{Classification+Regression} & $\mathbf{81.5}$   & $84.7$      & $\mathbf{83.1}$ & $0.19$     \\ \hline
\end{tabular}
\end{table}

\begin{table}[]
 \renewcommand\arraystretch{1.2}
 \centering
 \caption{Performance of different NMS types in the post-processing on CTW1500.}
 \label{tnms}
\begin{tabular}{c|cccc}
\hline
Type & Recall & Precision & F-measure & FPS   \\ \hline
NMS      & $81.8$   & $82.6$      & $82.2$      & $17.0$ \\
K-NMS    & $\mathbf{81.9}$   & $82.9$      & $82.4$      & $16.8$ \\
\textbf{S-NMS}    & $81.5$   & $\mathbf{84.7}$      & $\mathbf{83.1}$      & $\mathbf{17.2}$ \\ \hline
\end{tabular}
\end{table}

\subsubsection{Shared head convolutions}
Here we investigate the effect of sharing head convolutions. The results on CTW1500 dataset are shown in Table~\ref{sc}, in which ``gap'' refers to the absolute value of the difference between the classification score and the location score. The location score means the intersection over union between the predicted box and the ground-truth box.
In Table~\ref{sc}, we can see that only sharing the first four convolutions in the heads of the box regression and the mask regression branches has little effect on the performance. 
If further sharing the first four convolutions in the heads of both the classification and regression branches, the F-measure is improved from $82.6$ to $83.1$. This is mainly due to the $0.03$ reduction of ``gap", resulting in a higher-quality regression to the positive sample with the highest classification score.



\begin{table*}[]

\centering
 \caption{Comparisons with state-of-the-art works on CTW1500 and Total-Text. Ext denotes extra training data, where Syn, MLT, ArT, and MixT denote SynthText\cite{gupta2016synthetic}, ICDAR2017-MLT\cite{nayef2017icdar2017},ICDAR-ArT\cite{chng2019icdar2019}, and a mixed dataset composed of SynthCurve\cite{liu2020abcnet}, COCO-Text\cite{veit2016coco}, and ICDAR2019-MLT\cite{nayef2019icdar2019}, respectively. $\dag$ indicates evaluating the performance of Total-Text with the IOU@0.5\cite{liu2019curved}, and the default evaluation matrix of Total-Text is DetEval\cite{ch2017total}.}
\label{tabelctwtota} 
\begin{tabular}{cc|ccccc|ccccc}

\hline
\multirow{2}{*}{Method} &
\multirow{2}{*}{Paper} &
\multicolumn{5}{c}{CTW1500}                                                                         & \multicolumn{5}{c}{Total-Text}                                                                      \\ \cline{3-12} 
                                          &                          &
                                          \multicolumn{1}{c}{Ext} &\multicolumn{1}{c}{Recall} & \multicolumn{1}{c}{Precision} & \multicolumn{1}{c}{F-measure} & FPS  &
                                          \multicolumn{1}{c}{Ext} &\multicolumn{1}{c}{Recall} & \multicolumn{1}{c}{Precision} & \multicolumn{1}{c}{F-measure} &  FPS    \\ \hline
TextSnake\cite{long2018textsnake}               & ECCV'$18$                           & \multicolumn{1}{c}{Syn}                                              & \multicolumn{1}{c}{$\mathbf{85.3}$}   & \multicolumn{1}{c}{$67.9$}      & \multicolumn{1}{c}{$75.6$}      & -    
& \multicolumn{1}{c}{Syn}
& \multicolumn{1}{c}{$74.5$}   & \multicolumn{1}{c}{$82.7$}      & \multicolumn{1}{c}{$78.4$}      & -    \\ 
SegLink++\cite{tang2019seglink++}           & PR'$19$                                                                              &
\multicolumn{1}{c}{Syn}   &
\multicolumn{1}{c}{$79.8$}   & \multicolumn{1}{c}{$82.8$}      & \multicolumn{1}{c}{$81.3$}      & -    & 
\multicolumn{1}{c}{Syn}   &
\multicolumn{1}{c}{$80.9$}   & \multicolumn{1}{c}{$82.1$}      & \multicolumn{1}{c}{$81.5$}      & -    \\
TextField$\dag$\cite{xu2019textfield}         & TIP'$19$                                         & \multicolumn{1}{c}{Syn}                                      & \multicolumn{1}{c}{$79.8$}   & \multicolumn{1}{c}{$83.0$}      & \multicolumn{1}{c}{$81.4$}      & $6.0$    & 
\multicolumn{1}{c}{Syn}   &
\multicolumn{1}{c}{$79.9$}   & \multicolumn{1}{c}{$81.2$}      & \multicolumn{1}{c}{$80.6$}      & $6.0$    \\ 
MSR\cite{2019MSR}              & IJCAI'$19$                                                                          & 
\multicolumn{1}{c}{Syn}   &
\multicolumn{1}{c}{$78.3$}   & \multicolumn{1}{c}{$85.0$}      & \multicolumn{1}{c}{$81.5$}      & $4.3$    & 
\multicolumn{1}{c}{Syn}   &
\multicolumn{1}{c}{$74.8$}   & \multicolumn{1}{c}{$83.8$}      & \multicolumn{1}{c}{$79.0$}      & $4.3$    \\
LOMO\cite{2019Look}               & CVPR'$19$                                  & \multicolumn{1}{c}{Syn}                                       & \multicolumn{1}{c}{$69.6$}   & \multicolumn{1}{c}{$\mathbf{89.2}$}      & \multicolumn{1}{c}{$78.4$}      & $4.4$    & 
\multicolumn{1}{c}{Syn} &
\multicolumn{1}{c}{$75.7$}   & \multicolumn{1}{c}{$86.6$}      & \multicolumn{1}{c}{$81.6$}      & $4.4$    \\
PSENet-1s$\dag$\cite{wang2019Shape}               & CVPR'$19$                                                                          & 
\multicolumn{1}{c}{MLT}   &
\multicolumn{1}{c}{$79.7$}   & \multicolumn{1}{c}{$84.8$}      & \multicolumn{1}{c}{$82.2$}      & $3.9$    & 
\multicolumn{1}{c}{MLT}   &
\multicolumn{1}{c}{$78.0$}   & \multicolumn{1}{c}{$84.0$}      & \multicolumn{1}{c}{$80.9$}      & $3.9$    \\
ATRR\cite{wang2019arbitrary}              & CVPR'$19$                                                                          & 
\multicolumn{1}{c}{-}   &
\multicolumn{1}{c}{$80.2$}   & \multicolumn{1}{c}{$80.1$}      & \multicolumn{1}{c}{$80.1$}      & $10.0$    & 
\multicolumn{1}{c}{-}   &
\multicolumn{1}{c}{$76.2$}   & \multicolumn{1}{c}{$80.9$}      & \multicolumn{1}{c}{$78.5$}      & -    \\
CRAFT\cite{baek2019character}              & CVPR'$19$                                                                          & 
\multicolumn{1}{c}{Syn}   &
\multicolumn{1}{c}{$81.1$}   & \multicolumn{1}{c}{$86.0$}      & \multicolumn{1}{c}{$83.5$}      & -    & 
\multicolumn{1}{c}{Syn}   &
\multicolumn{1}{c}{$79.9$}   & \multicolumn{1}{c}{$87.6$}      & \multicolumn{1}{c}{$83.6$}      & -    \\
PAN\cite{wang2019efficient}              & ICCV'$19$                                                                          & 
\multicolumn{1}{c}{Syn}   &
\multicolumn{1}{c}{$81.2$}   & \multicolumn{1}{c}{$86.4$}      & \multicolumn{1}{c}{$83.7$}      & $\mathbf{39.8}$    & 
\multicolumn{1}{c}{Syn}   &
\multicolumn{1}{c}{$81.0$}   & \multicolumn{1}{c}{$89.3$}      & \multicolumn{1}{c}{$85.0$}      & $\mathbf{39.6}$    \\ 
Mask-TTD\cite{liu2019arbitrarily}               & TIP'$20$                                                                         & 

\multicolumn{1}{c}{-}   &
\multicolumn{1}{c}{$79.0$}   & \multicolumn{1}{c}{$79.7$}      & \multicolumn{1}{c}{$79.4$}      & -    & 
\multicolumn{1}{c}{-}   &
\multicolumn{1}{c}{$74.5$}   & \multicolumn{1}{c}{$79.1$}      & \multicolumn{1}{c}{$76.7$}      & -    \\
DBNet\cite{liao2020Real}              & AAAI'$20$                                                                          & 
\multicolumn{1}{c}{Syn}   &
\multicolumn{1}{c}{$80.2$}   & \multicolumn{1}{c}{$86.9$}      & \multicolumn{1}{c}{$83.4$}      & $22.0$    & 
\multicolumn{1}{c}{Syn}   &
\multicolumn{1}{c}{$82.5$}   & \multicolumn{1}{c}{$87.1$}      & \multicolumn{1}{c}{$84.7$}      & $32.0$    \\
TextRay$\dag$\cite{wang2020textray}              & MM'$20$                                                                         & 
\multicolumn{1}{c}{ArT}   &
\multicolumn{1}{c}{$80.4$}   & \multicolumn{1}{c}{$82.8$}      & \multicolumn{1}{c}{$81.6$}      & -    & 
\multicolumn{1}{c}{ArT}   &
\multicolumn{1}{c}{$77.9$}   & \multicolumn{1}{c}{$83.5$}      & \multicolumn{1}{c}{$80.6$}      & -    \\ 
ABCNet$\dag$\cite{liu2020abcnet}              & CVPR'$20$                                                                        & 
\multicolumn{1}{c}{MixT}   &
\multicolumn{1}{c}{$78.5$}   & \multicolumn{1}{c}{$84.4$}      & \multicolumn{1}{c}{$81.4$}      & -    & 
\multicolumn{1}{c}{MixT}   &
\multicolumn{1}{c}{$81.3$}   & \multicolumn{1}{c}{$87.9$}      & \multicolumn{1}{c}{$84.5$}      & -    \\ 
ContourNet\cite{wang2020contournet}              & CVPR'$20$                                                                          & 
\multicolumn{1}{c}{-}   &
\multicolumn{1}{c}{$84.1$}   & \multicolumn{1}{c}{$83.7$}      & \multicolumn{1}{c}{$83.9$}      & $4.5$    & 
\multicolumn{1}{c}{-}   &
\multicolumn{1}{c}{$83.9$}   & \multicolumn{1}{c}{$86.9$}      & \multicolumn{1}{c}{$85.4$}      & $3.8$    \\
DRRG\cite{2020Deep}              & CVPR'$20$                                                                          & 
\multicolumn{1}{c}{MLT}   &
\multicolumn{1}{c}{$83.0$}   & \multicolumn{1}{c}{$85.9$}      & \multicolumn{1}{c}{$84.5$}      & -    & 
\multicolumn{1}{c}{MLT}   &
\multicolumn{1}{c}{$\mathbf{84.9}$}   & \multicolumn{1}{c}{$86.5$}      & \multicolumn{1}{c}{$85.7$}      & -    \\
ReLaText\cite{ma2021relatext}              & PR'$21$                                                                          & 
\multicolumn{1}{c}{Syn}   &
\multicolumn{1}{c}{$83.3$}   & \multicolumn{1}{c}{$86.2$}      & \multicolumn{1}{c}{$84.8$}      & $10.6$    & 
\multicolumn{1}{c}{Syn}   &
\multicolumn{1}{c}{$83.1$}   & \multicolumn{1}{c}{$84.8$}      & \multicolumn{1}{c}{$84.0$}      & $3.2$    \\
OPMP\cite{zhang2020opmp}              & TMM'$21$                                                                       & 
\multicolumn{1}{c}{-}   &
\multicolumn{1}{c}{$80.8$}   & \multicolumn{1}{c}{$85.1$}      & \multicolumn{1}{c}{$82.9$}      & $1.4$    & 
\multicolumn{1}{c}{Syn}   &
\multicolumn{1}{c}{$82.7$}   & \multicolumn{1}{c}{$87.6$}      & \multicolumn{1}{c}{$85.1$}      & $1.4$    \\
Dai \etal\cite{dai2021accurate}  & TMM'$21$                                                                      & 
\multicolumn{1}{c}{-}   &
\multicolumn{1}{c}{$80.4$}   & \multicolumn{1}{c}{$86.2$}      & \multicolumn{1}{c}{$83.2$}      & $0.6$    & 
\multicolumn{1}{c}{-}   &
\multicolumn{1}{c}{$81.2$}   & \multicolumn{1}{c}{$85.4$}      & \multicolumn{1}{c}{$83.2$}      & $0.7$    \\
PCR\cite{dai2021progressive}              & CVPR'$21$                                                                          & 
\multicolumn{1}{c}{MLT}   &
\multicolumn{1}{c}{$82.3$}   & \multicolumn{1}{c}{$87.2$}      & \multicolumn{1}{c}{$84.7$}      & -    & 
\multicolumn{1}{c}{MLT}   &
\multicolumn{1}{c}{$82.0$}   & \multicolumn{1}{c}{$88.5$}      & \multicolumn{1}{c}{$85.2$}      & -    \\ 
FCENet\cite{zhu2021fourier}              & CVPR'$21$                                                                          & 
\multicolumn{1}{c}{-}   &
\multicolumn{1}{c}{$83.4$}   & \multicolumn{1}{c}{$87.6$}      & \multicolumn{1}{c}{$\mathbf{85.5}$}      & -    & 
\multicolumn{1}{c}{-}   &
\multicolumn{1}{c}{$82.5$}   & \multicolumn{1}{c}{$89.3$}      & \multicolumn{1}{c}{$85.8$}      & -    \\ 
TextBPN\cite{zhang2021adaptive}              & ICCV'$21$                                                                          & 
\multicolumn{1}{c}{Syn}   &
\multicolumn{1}{c}{$81.4$}   & \multicolumn{1}{c}{$87.8$}      & \multicolumn{1}{c}{$84.5$}      & $12.1$    & 
\multicolumn{1}{c}{Syn}   &
\multicolumn{1}{c}{$84.6$}   & \multicolumn{1}{c}{$\mathbf{90.2}$}      & \multicolumn{1}{c}{$\mathbf{87.3}$}      & 12.6    \\ \hline

\textbf{TextDCT}                 & Ours                                                                          & 
\multicolumn{1}{c}{-}   &
\multicolumn{1}{c}{$81.5$}   & \multicolumn{1}{c}{$84.7$}      & \multicolumn{1}{c}{$83.1$}      & $17.3$ & 
\multicolumn{1}{c}{-}   &
\multicolumn{1}{c}{$80.5$}   & \multicolumn{1}{c}{$85.8$}      & \multicolumn{1}{c}{$83.0$}      & $15.2$ \\ 
\textbf{TextDCT}                    & Ours                                                                       & 
\multicolumn{1}{c}{Syn}   &
\multicolumn{1}{c}{$\mathbf{85.3}$}   & \multicolumn{1}{c}{$85.0$}      & \multicolumn{1}{c}{$85.1$}      & $17.2$ & 
\multicolumn{1}{c}{Syn}   &
\multicolumn{1}{c}{$82.7$}   & \multicolumn{1}{c}{$87.2$}      & \multicolumn{1}{c}{$84.9$}      & $15.1$ \\ \hline 

\end{tabular}
\end{table*}

\begin{table*}[]
\centering
 \caption{Comparisons with state-of-the-art works on ICDAR2015 and MLT.}
 \label{tabelic157}
\centering
\begin{tabular}{cc|ccccc|ccccc}
\hline
\multirow{2}{*}{Method} &
\multirow{2}{*}{Paper}&
\multicolumn{5}{c}{ICDAR2015}                                                    & \multicolumn{5}{c}{MLT}                                                                      \\ \cline{3-12} 
                          &                    &  \multicolumn{1}{c}{Ext} &
                      \multicolumn{1}{c}{Recall} &  \multicolumn{1}{c}{Precision} &  \multicolumn{1}{c}{F-measure}  &
                     FPS  &
                 \multicolumn{1}{c}{Ext} &
                     \multicolumn{1}{c}{Recall}  & \multicolumn{1}{c}{Precision} & \multicolumn{1}{c}{F-measure} & FPS      \\ \hline

RRPN\cite{ma2018arbitrary}                 & TMM'$18$                                                                         & \multicolumn{1}{c}{-}   & \multicolumn{1}{c}{$73.0$}      &
\multicolumn{1}{c}{$82.0$}      &
\multicolumn{1}{c}{$77.0$}      & -
& \multicolumn{1}{c}{-} & \multicolumn{1}{c}{$55.5$}   & \multicolumn{1}{c}{$71.2$}      & \multicolumn{1}{c}{$62.4$}      & - \\ 

Lyu \etal\cite{lyu2018multi}                 & CVPR'$18$                                                                         & \multicolumn{1}{c}{Syn}   & \multicolumn{1}{c}{$70.7$}      &
\multicolumn{1}{c}{$\mathbf{94.1}$}      &
\multicolumn{1}{c}{$80.7$}      & $3.6$
& \multicolumn{1}{c}{Syn} & \multicolumn{1}{c}{$55.6$}   & \multicolumn{1}{c}{$83.8$}      & \multicolumn{1}{c}{$66.8$}      & $5.7$ \\ 
SPCNet\cite{xie2019scene}                 & AAAI'$19$                                                                         & \multicolumn{1}{c}{MLT}   & \multicolumn{1}{c}{$85.8$}      &
\multicolumn{1}{c}{$88.7$}      &
\multicolumn{1}{c}{$87.2$}      & -
& \multicolumn{1}{c}{Syn} & \multicolumn{1}{c}{$\mathbf{73.4}$}   & \multicolumn{1}{c}{$76.9$}      & \multicolumn{1}{c}{$70.0$}      & - \\
ATRR\cite{wang2019arbitrary}              & CVPR'$19$                                                                          & 
\multicolumn{1}{c}{-}   &
\multicolumn{1}{c}{$86.0$}   & \multicolumn{1}{c}{$89.2$}      & \multicolumn{1}{c}{$87.6$}      & -    & 
\multicolumn{1}{c}{-}   &
\multicolumn{1}{c}{-}   & \multicolumn{1}{c}{-}      & \multicolumn{1}{c}{-}      & -    \\

PSENet-1s\cite{tang2019seglink++}                 & CVPR'$19$                                                                         & \multicolumn{1}{c}{MLT}   & \multicolumn{1}{c}{$84.5$}      &
\multicolumn{1}{c}{$86.9$}      &
\multicolumn{1}{c}{$85.7$}      & $1.6$
& \multicolumn{1}{c}{-} & \multicolumn{1}{c}{$68.2$}   & \multicolumn{1}{c}{$73.7$}      & \multicolumn{1}{c}{$70.8$}      & - \\

LOMO\cite{2019Look}                 & CVPR'$19$                                                                         & \multicolumn{1}{c}{Syn}   & \multicolumn{1}{c}{$83.5$}      &
\multicolumn{1}{c}{$91.3$}      &
\multicolumn{1}{c}{$87.2$}      & $3.4$
& \multicolumn{1}{c}{Syn} & \multicolumn{1}{c}{$60.6$}   & \multicolumn{1}{c}{$78.8$}      & \multicolumn{1}{c}{$68.5$}      & - \\

CRAFT\cite{baek2019character}                 & CVPR'$19$                                                                         & \multicolumn{1}{c}{Syn}   & \multicolumn{1}{c}{$84.3$}      &
\multicolumn{1}{c}{$89.8$}      &
\multicolumn{1}{c}{$86.9$}      & $8.6$
& \multicolumn{1}{c}{Syn} & \multicolumn{1}{c}{$68.2$}   & \multicolumn{1}{c}{$80.6$}      & \multicolumn{1}{c}{$73.9$}      & $8.6$ \\

DBNet\cite{liao2020Real}              & AAAI'$20$                                                                          & \multicolumn{1}{c}{Syn}   & \multicolumn{1}{c}{$83.2$}      &
\multicolumn{1}{c}{$91.8$}      &
\multicolumn{1}{c}{$87.3$}      & $12$
& \multicolumn{1}{c}{Syn} & \multicolumn{1}{c}{$67.9$}   & \multicolumn{1}{c}{$83.1$}      & \multicolumn{1}{c}{$74.7$}      & $\mathbf{19}$ \\

DRRG\cite{2020Deep}                 & CVPR'$20$                                                                         & \multicolumn{1}{c}{MLT}   & \multicolumn{1}{c}{$84.7$}      &
\multicolumn{1}{c}{$88.5$}      &
\multicolumn{1}{c}{$86.6$}      & -
& \multicolumn{1}{c}{-} & \multicolumn{1}{c}{$61.0$}   & \multicolumn{1}{c}{$75.0$}      & \multicolumn{1}{c}{$67.3$}      & - \\

ContourNet\cite{wang2020contournet}              & CVPR'$20$                                                                        & \multicolumn{1}{c}{-}   & \multicolumn{1}{c}{$86.1$}      &
\multicolumn{1}{c}{$87.6$}      &
\multicolumn{1}{c}{$86.9$}      & $3.5$
& \multicolumn{1}{c}{-} & \multicolumn{1}{c}{-}   & \multicolumn{1}{c}{-}      & \multicolumn{1}{c}{-}      & - \\

Boundary\cite{xing2021boundary}                 & MM'$20$                                                                         & \multicolumn{1}{c}{Syn}   & \multicolumn{1}{c}{$82.2$}      &
\multicolumn{1}{c}{$88.1$}      &
\multicolumn{1}{c}{$85.0$}      & -
& \multicolumn{1}{c}{-} & \multicolumn{1}{c}{-}   & \multicolumn{1}{c}{-}      & \multicolumn{1}{c}{-}      & - \\

MS-CAFA\cite{dai2019deep}                 & TMM'$20$                                                                         & \multicolumn{1}{c}{-}   & \multicolumn{1}{c}{$82.7$}      &
\multicolumn{1}{c}{$86.2$}      &
\multicolumn{1}{c}{$84.4$}      & $0.5$
& \multicolumn{1}{c}{-} & \multicolumn{1}{c}{$66.8$}   & \multicolumn{1}{c}{$79.5$}      & \multicolumn{1}{c}{$72.6$}      & $0.5$ \\

R-Net\cite{wang2020r}                 & TMM'$21$                                                                         & \multicolumn{1}{c}{Syn}   & \multicolumn{1}{c}{$82.8$}      &
\multicolumn{1}{c}{$88.7$}      &
\multicolumn{1}{c}{$85.6$}      & $\mathbf{21.4}$
& \multicolumn{1}{c}{Syn} & \multicolumn{1}{c}{$64.5$}   & \multicolumn{1}{c}{$70.4$}      & \multicolumn{1}{c}{$67.3$}      & - \\

TextMountain\cite{zhu2021textmountain}                 & PR'$21$                                                                         & \multicolumn{1}{c}{Syn}   & \multicolumn{1}{c}{$84.1$}      &
\multicolumn{1}{c}{$87.3$}      &
\multicolumn{1}{c}{$85.7$}      & $10.4$
& \multicolumn{1}{c}{-} & \multicolumn{1}{c}{$55.2$}   & \multicolumn{1}{c}{$80.8$}      & \multicolumn{1}{c}{$65.6$}      & $6.0$ \\

FCENet\cite{zhu2021fourier}                 & CVPR'$21$                                                                         & \multicolumn{1}{c}{-}   & \multicolumn{1}{c}{$82.6$}      &
\multicolumn{1}{c}{$90.1$}      &
\multicolumn{1}{c}{$86.2$}      & -
& \multicolumn{1}{c}{-} & \multicolumn{1}{c}{-}   & \multicolumn{1}{c}{-}      & \multicolumn{1}{c}{-}      & - \\

MOST\cite{he2021most}                 & CVPR'$21$                                                                         & \multicolumn{1}{c}{Syn}   & \multicolumn{1}{c}{$\mathbf{87.3}$}      &
\multicolumn{1}{c}{$89.1$}      &
\multicolumn{1}{c}{$\mathbf{88.2}$}      & $10.0$
& \multicolumn{1}{c}{Syn} & \multicolumn{1}{c}{$72.0$}   & \multicolumn{1}{c}{$82.0$}      & \multicolumn{1}{c}{$\mathbf{76.7}$}      & $10.1$ \\\hline

\textbf{TextDCT}                 & Ours                                                                         & \multicolumn{1}{c}{-}   & \multicolumn{1}{c}{$83.7$}      &
\multicolumn{1}{c}{$86.9$}      &
\multicolumn{1}{c}{$85.3$}      & $7.6$
& \multicolumn{1}{c}{-} & \multicolumn{1}{c}{$68.0$}   & \multicolumn{1}{c}{$79.0$}      & \multicolumn{1}{c}{$73.3$}      & $11.6$ \\

\textbf{TextDCT}                 & Ours                                                                         & \multicolumn{1}{c}{Syn}   & \multicolumn{1}{c}{$84.8$}      &
\multicolumn{1}{c}{$88.9$}      &
\multicolumn{1}{c}{$86.8$}      & $7.5$
& \multicolumn{1}{c}{Syn} & \multicolumn{1}{c}{$67.7$}   & \multicolumn{1}{c}{$\mathbf{83.9}$}      & \multicolumn{1}{c}{$74.9$}      & $11.8$ \\
\hline

\end{tabular}
\end{table*}

\subsubsection{S-NMS}
In Table~\ref{tnms}, we filter the redundant boxes using NMS, S-NMS, and kernel-level NMS (K-NMS) in our TextDCT, respectively. K-NMS is a two-stage NMS, which first uses NMS for each text kernel corresponding to the regression box separately, and then uses NMS for the remaining boxes. Compared with NMS, K-NMS has a slightly different filtering order, resulting in slightly different final results ($82.4$ vs. $82.2$ in terms of F-measure). 
Compared with NMS and KNMS, S-NMS can effectively improve the performance. 
Note that there may be adhesions between the predicted adjacent text kernels. Since S-NMS only selects the text box corresponding to the highest classification score for the adhesion region, the recall is decreased slightly. Although with $0.3$ recall rate drops on the CTW1500, S-NMS exceeds NMS by $0.9$ in terms of F-measure, in which F-measure gives a convincing measurement result by balancing precision and recall.


\subsection{Comparison with State-of-the-Art Methods}


For a fair comparison, we only evaluate our model on a single scale for all datasets. 
Furthermore, for the text spotting model\cite{liu2020abcnet,liu2018fots,lyu2018mask}, we only show the detection results without recognition branch. 


\subsubsection{Evaluation on Long Curved Text Benchmark}
The comparison results on long curved text dataset CTW1500 are given in Table~\ref{tabelctwtota}, from which we can see that TextDCT achieves competitive performance and speed. 
Without pre-training, TextDCT can still achieve competitive results with $81.5$, $84.7$, and $83.1$ in recall, precision, and F-measure, respectively.

On one hand, TextDCT outperforms at least $0.6$ than segmentation based methods\cite{xu2019textfield,liao2020Real,wang2019efficient,wang2019Shape,zhang2021adaptive} in F-measure, which are more sensitive to noise and require pre-training on additional datasets to achieve good results. For example, the F-measure of PSENet\cite{wang2019Shape} after pre-training on the Synthtext\cite{gupta2016synthetic} dataset is $4.2$ higher than that without pre-training. 
Although the precision of DBNet\cite{liao2020Real} is $1.9$ higher than TextDCT, some small texts may be filtered out in the post-processing of DBNet, so its recall rate is much lower than our TextDCT ($80.2$ vs. $85.3$).

On the other hand, for regression based methods, EAST\cite{zhou2017east} only regresses quadrilaterals so it cannot adapt to curved texts ($60.4$ in F-measure). Although LOMO\cite{2019Look} strives to achieve accurate text region representation by iterative optimization module, TextDCT obtains more higher performance ($85.1$ vs. $80.8$). Moreover, compared with FCENet\cite{zhu2021fourier}, TextDCT is more capable of modeling extremely long texts, which will be explored in detail in Sec.~\ref{ssec:dct_dft}.

\begin{figure*}[!ht]
    \centering
    \setlength{\abovecaptionskip}{0cm}
    \setlength{\belowcaptionskip}{0cm}
    \subfigure[CTW1500]
    {
        \includegraphics[width=\textwidth,height=3cm]{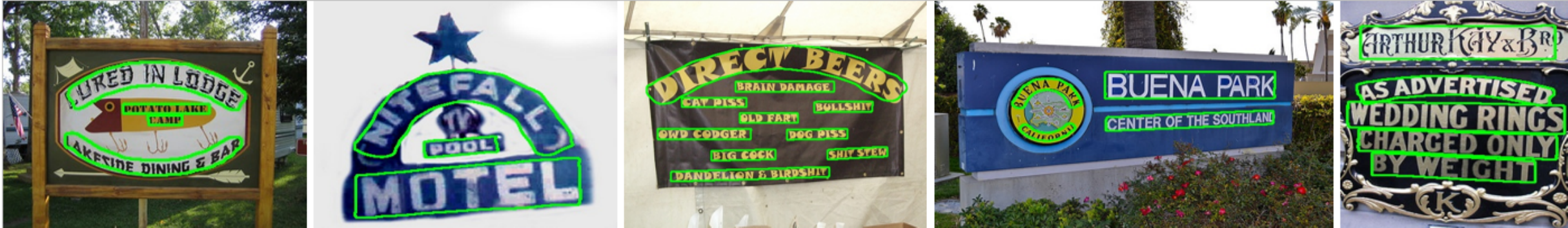}
        \label{fig8a}
    }
    \subfigure[Total-Text]
    {
        \includegraphics[width=\textwidth,height=3cm]{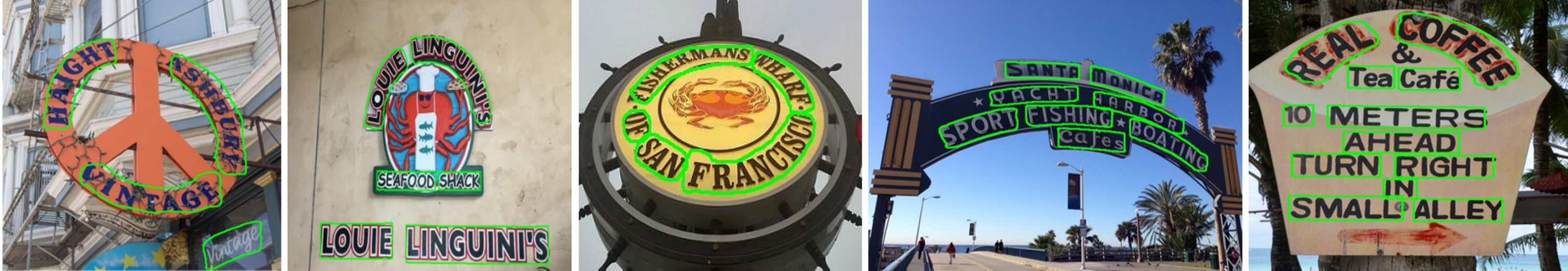}
        \label{fig8b}
    }
    \subfigure[ICDAR2015]
    {
        \includegraphics[width=\textwidth,height=3cm]{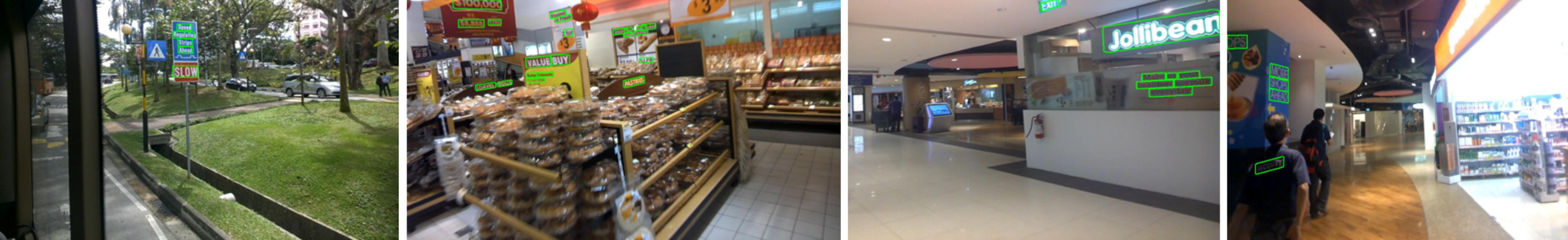}
        \label{fig8c}
    }
    \subfigure[MLT]
    {
        \includegraphics[width=\textwidth,height=3cm]{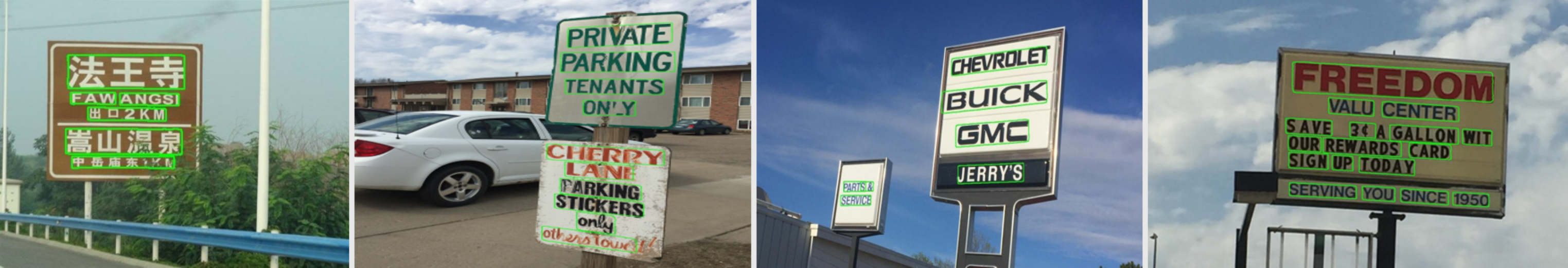}
        \label{fig8d}
    }
    \caption{Example results of our TextDCT on scene text datasets CTW1500, Total-Text, ICDAR2015, and MLT.}
    \label{fig8}
   
\end{figure*}

\subsubsection{Evaluation on Curved Text Benchmark}
The comparison results on curved text dataset Total-Text are shown in Table~\ref{tabelctwtota}, where we evaluate using the protocol in\cite{ch2017total}.

We can observe that TextDCT in a single testing scale achieves competitive performance and speed ($84.9$ in F-measure and 15.1 in FPS). As a single-stage regression based method, TextDCT significantly outperforms existing regression based methods\cite{lyu2018mask,2019Look,wang2020textray}. 
As a two-stage method, TextSpotter\cite{lyu2018mask} performs end-to-end text detection and recognition by modifying the mask branch of Mask-RCNN\cite{he2017mask},  which has much lower detection capability than the single-stage TextDCT ($84.4$ vs. $78.5$ in F-measure). Compared with SPCNet\cite{xie2019scene}, which is a two-stage regression based model with a text context module (TCM) and a re-score mechanism to improve model performance, TextDCT outperforms it by $4.2$ and $2.0$ in recall and F-measure respectively. Besides, as a single-stage model with a single-level head, TextDCT has a much simpler pipeline than these two-stage models. CRAFT\cite{baek2019character} additionally uses character-level annotations to supervise the learning process, which greatly increases the difficulty of annotating datasets. Compared to CRAFT, our TextDCT trains with only word-level supervision and shows advantages in F-measure.

Note that TextRay\cite{wang2020textray} models text contours in polar coordinates and represents text contours by a finite number of contour points, which may have limited ability to model texts with highly curved shapes (see Fig.~\ref{fig1}). However, TextDCT can model extremely curved texts well with DCT mask representation and achieve better results. 
Compared with \cite{dai2019deep} improving the discrimination of text feature representations through the multi-scale context aware feature aggregation module, TextDCT achieves much better performance ($84.9$ vs. $81.5$ in F-measure) by our single-level framework and DCT mask representation. Besides, compared with the segmentation based methods\cite{long2018textsnake,xu2019textfield} that include complex post-processing and are susceptible to false positives, our TextDCT uses a simple pipeline and suppresses false positives through S-NMS, outperforming these methods by a large margin. 

Some qualitative results on Total-Text dataset are depicted in Fig.~\ref{fig8b}. As a light-weighted single-shot regression-based detector, our TextDCT achieves satisfactory results in various degrees of curvature, aspect ratios and complex backgrounds. 

\begin{figure*}[!ht]
    \centering
    \subfigure
    {
        \includegraphics[width=\textwidth,height=5.8cm]{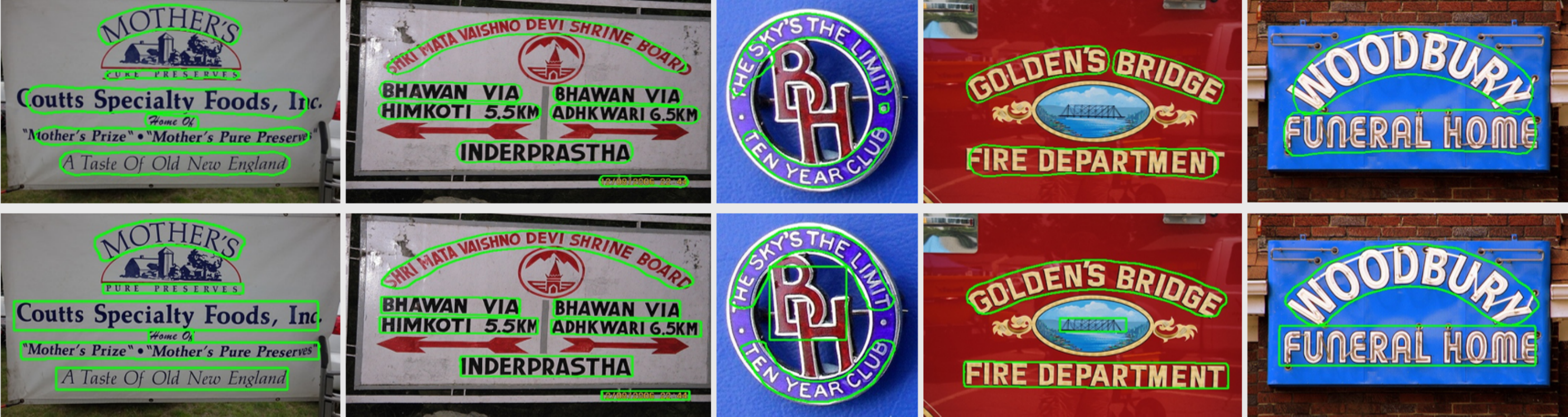}

    }

    \caption{Qualitative comparisons on challenging example images in CTW1500. The top and bottom rows show the results of FCENet* and our TextDCT, respectively.}
    \label{figdctabc}
\end{figure*}

\subsubsection{Evaluation on Oriented Text Benchmark}
Test results on ICDAR2015 following the standard evaluation metric are shown in Table~\ref{tabelic157}. We can see that our TextDCT achieves $83.7$, $86.9$, and $85.3$ for recall, precision, and F-measure respectively without any extra datasets, and achieves satisfactory results in terms of F-measure ($86.8$ in F-measure) when pre-trained on SynthText.

As shown in Table~\ref{tabelic157}, TextDCT surpasses most existing regression based methods\cite{zhou2017east,dai2019deep,dai2021accurate,wang2020r}. 
Specifically, compared with EAST\cite{zhou2017east}, although both are single-stage models, our model can achieve spatial-awareness and scale-awareness through FAM, and has better detection capabilities for long texts in ICDAR2015 ($86.8$ vs. $80.7$). R-Net\cite{wang2020r} achieves spatial-awareness and scale-awareness by a spatial relationship module (SPM) and a scale relationship module (SRM). However, TextDCT can achieve better performance since more accurate contours can be obtained by mask regression branch and S-NMS eliminates some false positives. Compared with the two-stage anchor-based methods\cite{dai2021accurate,dai2019deep}, in which the latter contains a more complex network structure and sophisticated post-processing, and is sensitive to the anchor setting with more hyper-parameters, TextDCT achieves much better performance. 
Some example results on ICDAR2015 are shown in Fig.~\ref{fig8c}.

\subsubsection{Evaluation on Multi-Lingual Benchmark}
The comparison results on MLT dataset for evaluating the ability of multi-lingual text detection are presented in Table~\ref{tabelic157}. 
Our TextDCT surpasses most existing methods such as LOMO\cite{2019Look}, R-Net\cite{wang2020r}, and TextMoutain\cite{zhu2021textmountain}. 
Besides, the example results on MLT are shown in Fig.~\ref{fig8d}, which shows that our TextDCT can effectively detect multi-lingual texts.

\begin{table}[]
\centering
 \caption{Quantitative comparison of F-measure results between FCENet* and TextDCT under different IOU thresholds on the challenging subset of CTW1500.}
 \label{fcedct}
 \renewcommand\arraystretch{1.2}
 
\begin{tabular}{c|cccc}
\hline
Method      & IOU@0.5  & IOU@0.6  & IOU@0.7                      & IOU@0.8                      \\ \hline
FCENet* & $80.9$ & 77.0 & $69.1$ & $46.8$ \\
TextDCT                      & $80.6$ & $77.2$ & $69.6$                     & $47.8$                     \\ \hline
\end{tabular}
\end{table}

\subsection{DCT Mask vs. DFT Contour}
\label{ssec:dct_dft}

Recently, there is a work FCENet\cite{zhu2021fourier} which adopts DFT to encode text contours as compact vectors. To fairly compare our DCT mask representation with the DFT contour representation, we implement FCENet* by applying the idea of FCENet to our TextDCT framework. The only difference between FCENet* and TextDCT is that FCENet* uses a classification branch and a contour regression branch in the single-level head, in which the box regression branch is discarded. 

Considering extremely long texts and extremely curved texts are the main challenges in scene text detection, we build a challenging subset with totally $347$ samples of CTW1500, in which the text whose text mask area is less than half of the text box area or the longest edge of the text box is longer than three-quarters of the longest edge of the image are selected.
Qualitative and quantitative comparisons were shown in Fig.~\ref{figdctabc} and Table.~\ref{fcedct}, respectively. 
Compared to our TextDCT, FCENet* tends to miss some corner pixels of long texts. Although the F-measure of FCENet* is $0.3$ higher than TextDCT under the evaluation protocol IOU@0.5, the F-measure of FCENet* drops more when the IOU threshold is increased from $0.5$ to $0.8$. The F-measure of TextDCT is $1.0$ higher than that of FCENet* under the evaluation protocol IOU@0.8, mainly because TextDCT fits text contours more accurately than FCENet*.

\subsection{Running Time Analysis}
In the inference stage, the running time of our TextDCT includes network inference time and post-processing time, where post-processing mainly lies in S-NMS and IDCT. We set different input sizes for different datasets to achieve optimal model performance. The size of input images and the number of text instances has a great impact on the model running time, as shown in Table~\ref{ttime}. We compute the average running time of each image in each dataset on an NVIDIA Tesla V100 GPU. The results reported in Table~\ref{ttime} show that our TextDCT is able to detect arbitrary-shaped scene texts quickly.  

\begin{table}[]
 \centering
 \caption{Running time of our TextDCT.}
 \label{ttime}
\begin{tabular}{cc|cc}
\hline
Dataset    & Input size & Network Inference & Post-processing  \\ \hline
CTW1500    & $1,056\times1,440$ & $0.022$s              & $0.036$s           \\
Total-Text & $1,024\times2,000$ & $0.028$s              & $0.038$s            \\
ICDAR-2015 & $1,500\times2,600$ & $0.054$s              & $0.077$s            \\
MLT        & $1,600\times2,000$ & $0.034$s              & $0.052$s         \\  \hline
\end{tabular}
\end{table}

\subsection{Limitations}
\begin{figure}[t]
    \centering
    \subfigure[]
    {
        \includegraphics[width=0.22\textwidth,height=2.8cm]{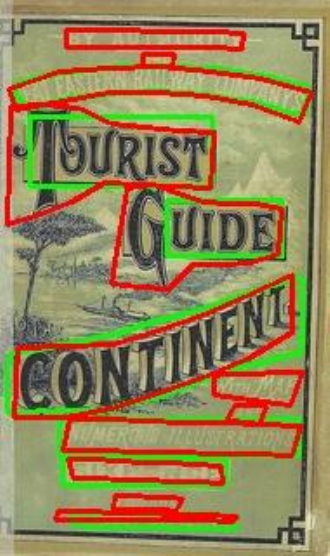}
    }
    \subfigure[]
    {
        \includegraphics[width=0.22\textwidth,height=2.8cm]{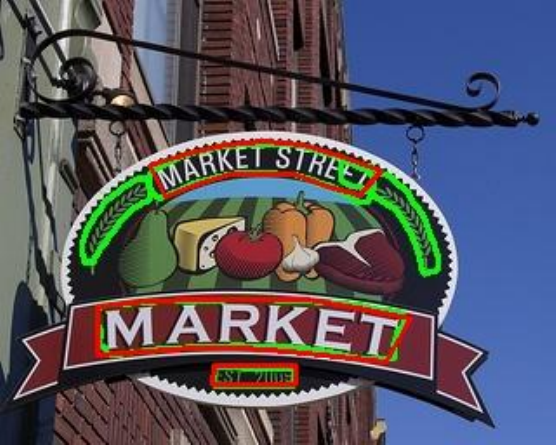}

    }

    \caption{Failure cases. Red contours are ground-truths while green contours are predicted results.}
    \label{fig9}
\end{figure}

According to the above experimental results, our TextDCT can perform well in most challenging scenarios, accurately detecting highly curved texts. However, there are a few failure cases, as shown in Fig.~\ref{fig9}. If the predicted text kernels stick to each other, it will cause omission because S-NMS only selects the point with the highest classification score for each kernel. Besides, some very text-like contexts are not filtered by our model, which should be alleviated by hard example mining.

\section{Conclusion}
In this paper, we have proposed a novel arbitrary-shaped scene text detector named TextDCT. The geometric encodings can be effectively learned through FAM and TKS, and false positives can be effectively removed by S-NMS. To obtain high-quality and low-complexity text instances, we have applied DCT to encode text instances as compact vectors. Extensive experiments have shown that the proposed single-level prediction framework can effectively detect arbitrary-shaped scene texts.

In the future work, the idea of mitigating the visual-semantic gap in \cite{yan2020semantics} can be explored to suppress the false positives like the failure cases in Fig.~\ref{fig9}. For example, we can extend TextDCT to an end-to-end scene text spotting framework, in which the text recognition module is trained to recognize texts as well as distinguish texts and text-like backgrounds. In this way, the detected text-like backgrounds can be removed based on the outputs of the text recognition module. Besides, our proposed idea of employing DCT to encode text masks as compact vectors is also promising to be applied for other object detection tasks.

\ifCLASSOPTIONcaptionsoff 
  \newpage
\fi

\bibliographystyle{IEEEtran}
\bibliography{IEEEabrv,reference}

\begin{IEEEbiography}[{\includegraphics[width=1in,height=1.25in,clip,keepaspectratio]{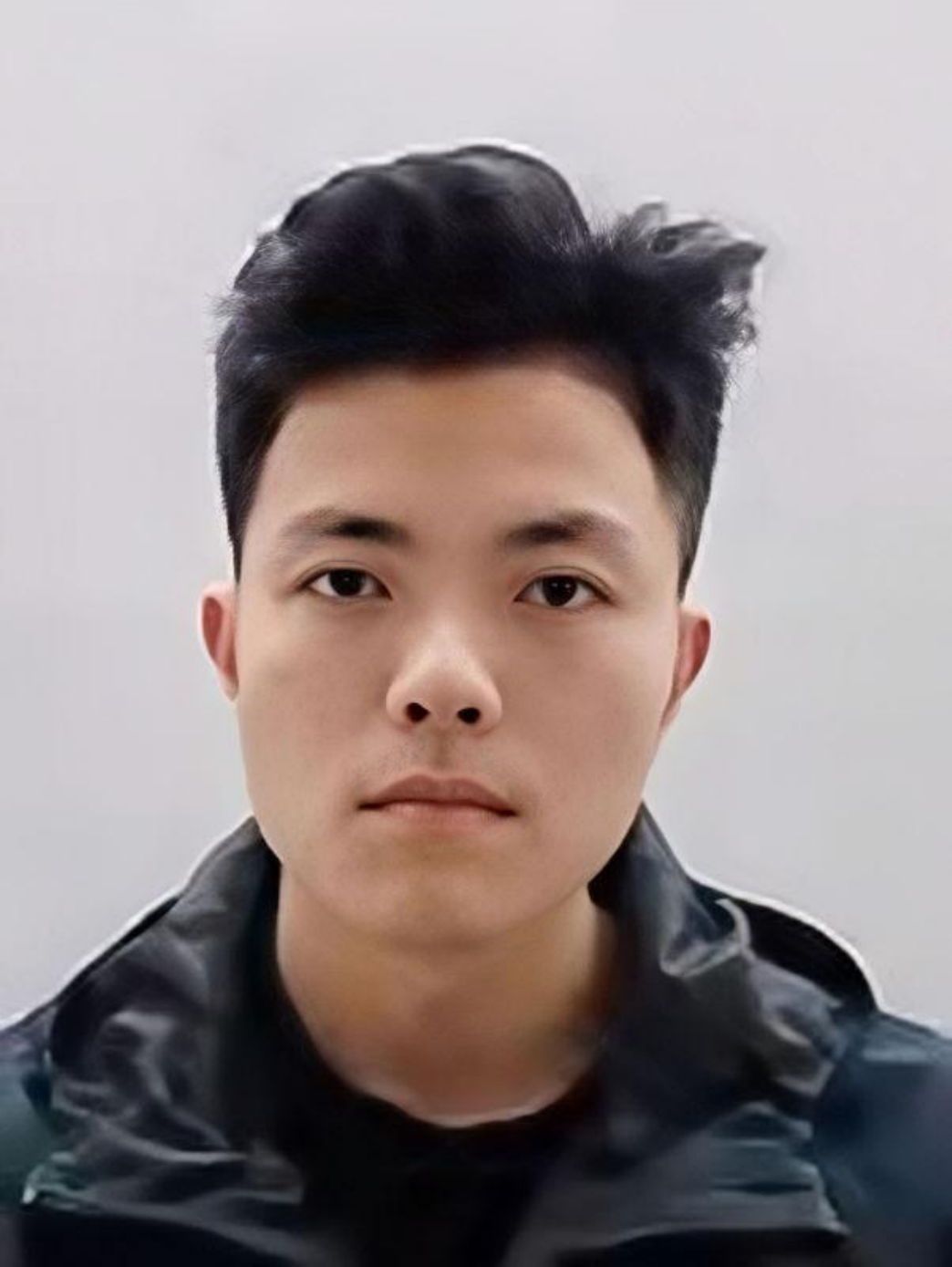}}]{Yuchen Su} is currently pursuing the M.S. degree at the School of Computer Science and Technology, China University of Mining and Technology, China, under the supervision of Prof. Yong Zhou, Prof. Fanrong Meng, and Dr. Zhiwen Shao. His current research interests include scene text detection and recognition. 
\end{IEEEbiography}

\begin{IEEEbiography}[{\includegraphics[width=1in,height=1.25in,clip,keepaspectratio]{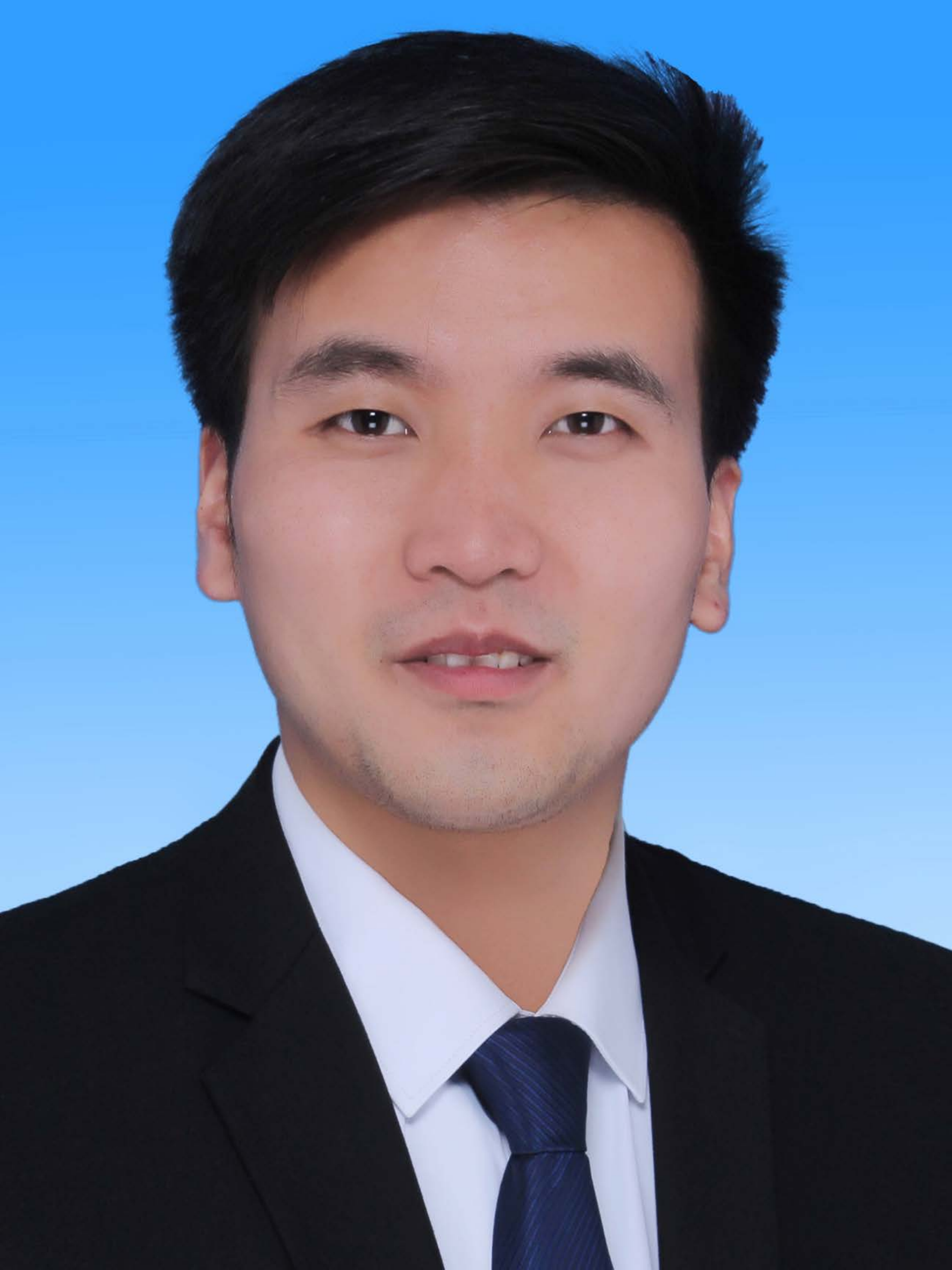}}]{Zhiwen Shao}
received his B.Eng. degree in Computer Science and Technology from the Northwestern Polytechnical University, China in 2015. He received the Ph.D. degree from the Shanghai Jiao Tong University, China in 2020. He is now a Tenure-Track Associate Professor at the School of Computer Science and Technology, China University of Mining and Technology, China. From 2017 to 2018, he was a joint Ph.D. student at the Multimedia and Interactive Computing Lab, Nanyang Technological University, Singapore. His research interests lie in computer vision and deep learning. He has been serving as a PC member in IJCAI and AAAI.
\end{IEEEbiography}

\begin{IEEEbiography}[{\includegraphics[width=1in,height=1.25in,clip,keepaspectratio]{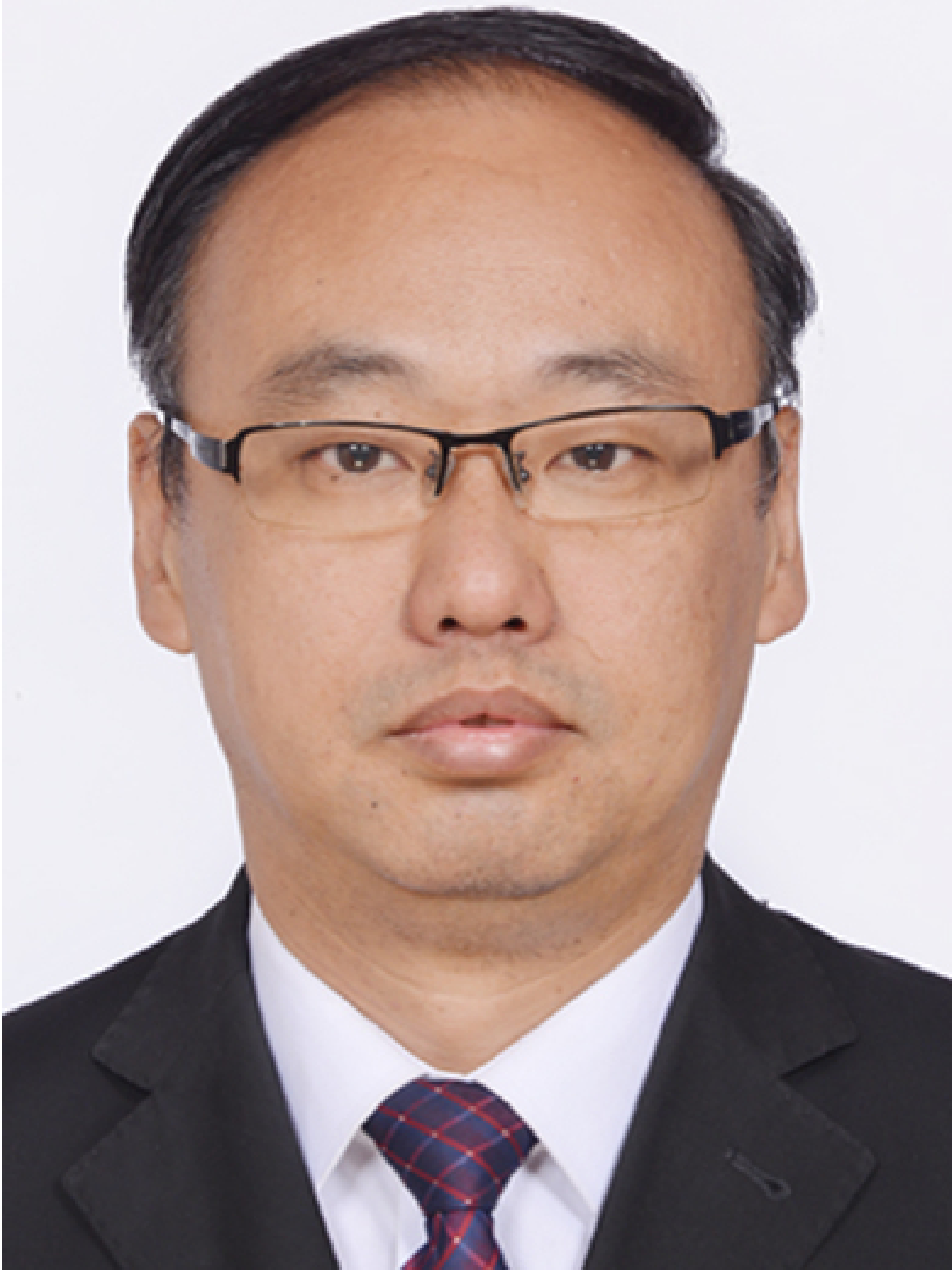}}]{Yong Zhou} received the M.S. and Ph.D. degrees in Control Theory and Control Engineering from the China University of Mining and Technology, China in 2003 and 2006, respectively. He is currently a Professor with the School of Computer Science and Technology, China University of Mining and Technology, China. His research interests include machine learning, intelligence
optimization, and data mining.
\end{IEEEbiography}

\begin{IEEEbiography}[{\includegraphics[width=1in,height=1.25in,clip,keepaspectratio]{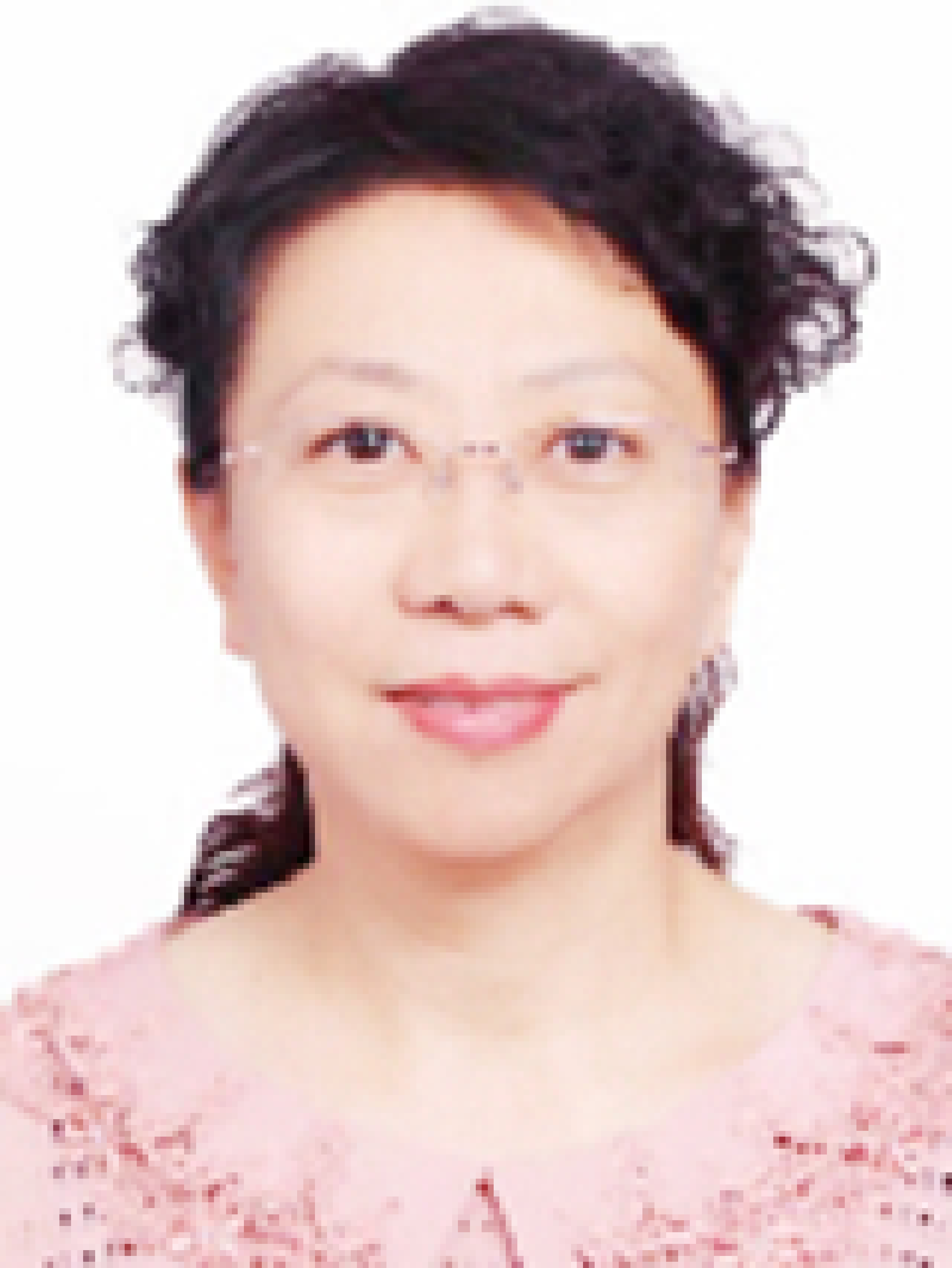}}]{Fanrong Meng} received the Ph.D. degree from the China University of Mining and Technology, China. She is currently a Professor with the School of Computer Science and Technology, China University of Mining and Technology, China. Her research interests include intelligent information processing, database technology, and data mining.
\end{IEEEbiography}

\begin{IEEEbiography}[{\includegraphics[width=1in,height=1.25in,clip,keepaspectratio]{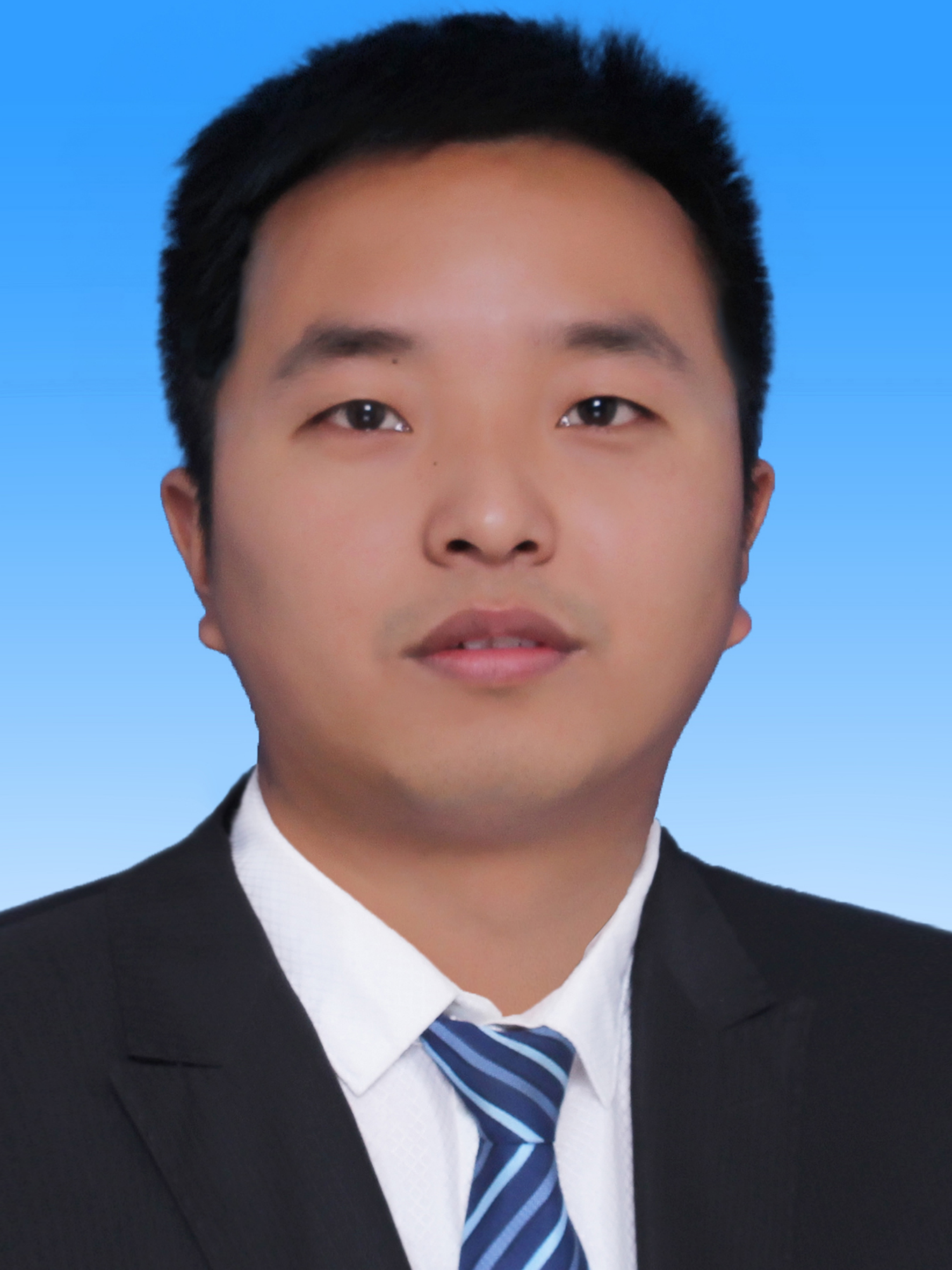}}]{Hancheng Zhu} received the B.S. degree from the Changzhou Institute of Technology, Changzhou, China, in 2012, and the M.S. and Ph.D. degrees from the China University of Mining and Technology, Xuzhou, China, in 2015 and 2020, respectively. He is currently a Postdoctoral Fellow at the School of Computer Science and Technology, China University of Mining and Technology, China. His research interests include image aesthetics assessment and affective computing.
\end{IEEEbiography}

\begin{IEEEbiography}[{\includegraphics[width=1in,height=1.25in,clip,keepaspectratio]{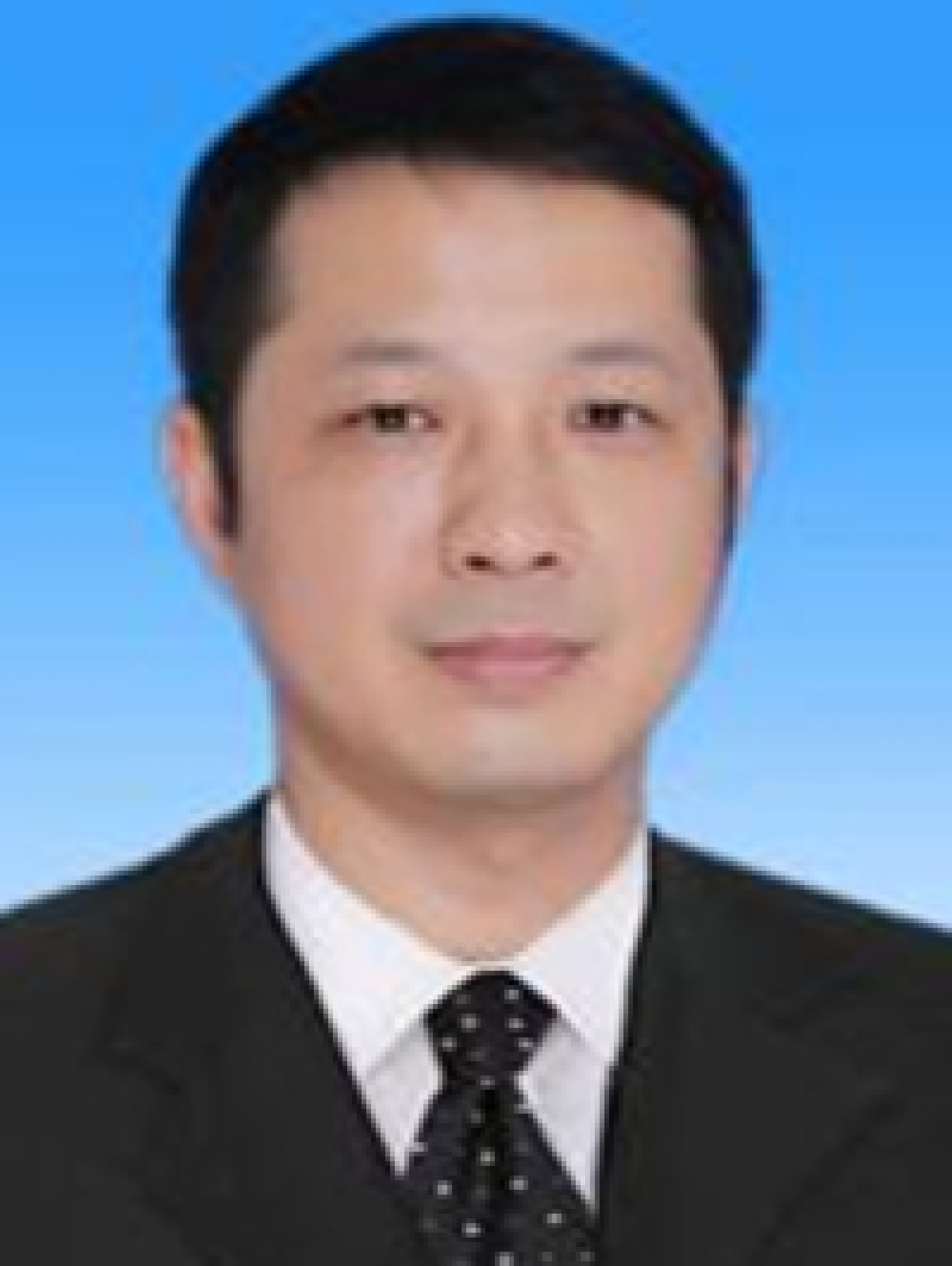}}]{Bing Liu} received the B.S., M.S., and Ph.D. degrees in 2002, 2005, and 2013, respectively, from the China University of Mining and Technology, Xuzhou, China. He is currently an Associate Professor at the School of Computer Science and Technology, China University of Mining and Technology, China. His current research interests include natural language processing, image understanding, and deep learning.
\end{IEEEbiography}

\begin{IEEEbiography}[{\includegraphics[width=1in,height=1.25in,clip,keepaspectratio]{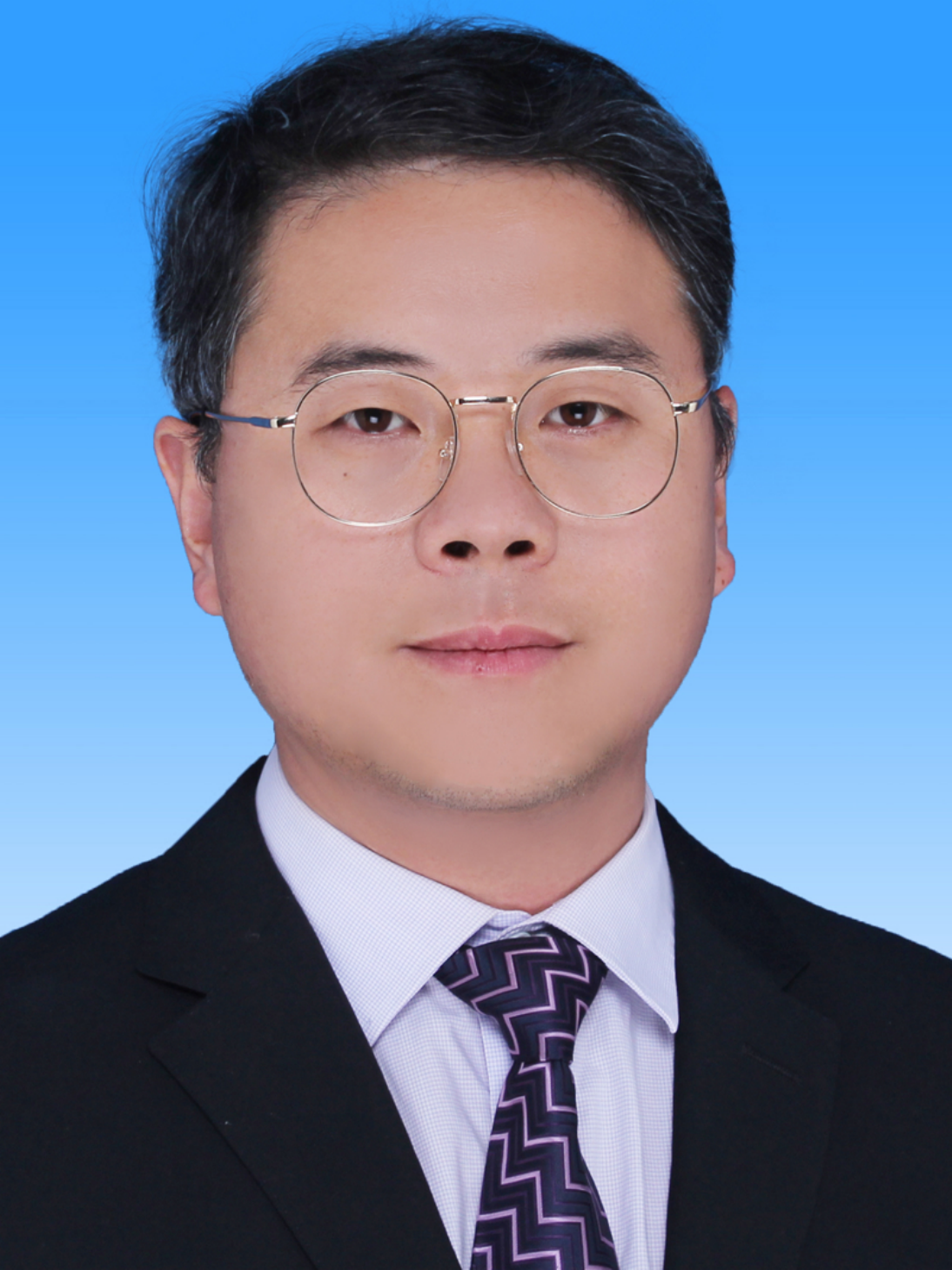}}]{Rui Yao} received the Ph.D. degree in computer science from the Northwestern Polytechnical University, Xi'an, China, in 2013. From 2011 to 2012, he was a Visiting Student with the University of Adelaide, Adelaide, SA, Australia. He is currently with the School of Computer Science and Technology, China University of Mining and Technology, Xuzhou, China. His research interests include computer vision and machine learning.
\end{IEEEbiography}

\end{document}